\pdfoutput=1
\documentclass[10pt]{article} 
\usepackage[preprint]{tmlr}

\usepackage{amsmath,amsfonts,bm}

\def\eqref#1{equation~\ref{#1}}

\def\1{\bm{1}}

\DeclareMathAlphabet{\mathsfit}{\encodingdefault}{\sfdefault}{m}{sl}
\SetMathAlphabet{\mathsfit}{bold}{\encodingdefault}{\sfdefault}{bx}{n}

\DeclareMathOperator*{\argmax}{arg\,max}

\usepackage{hyperref}
\usepackage{url}

\usepackage{graphicx} 
\usepackage{subcaption}
\usepackage[english]{babel}
\usepackage[utf8x]{inputenc}
\usepackage[T1]{fontenc}
\usepackage{float}
\usepackage{xurl}
\usepackage{hyperref}
\usepackage{footmisc}
\usepackage{color}
\usepackage{tabularx}
\usepackage{xtab,booktabs}
\usepackage{caption}
\usepackage{algpseudocode}
\usepackage{etoolbox}
\usepackage{enumitem}
\usepackage{tabularx}
\usepackage{pifont}
\usepackage{multirow}
\usepackage{relsize}
\usepackage{makecell}
\usepackage{float}
\usepackage{enumitem}
\usepackage{algorithm}
\usepackage{algpseudocode}
\usepackage{amsmath}
\usepackage{amssymb}
\usepackage{amsthm}
\usepackage{algpseudocodex}
\usepackage{xcolor}
\usepackage{colortbl}

\DeclareMathOperator*{\expect}{{\Large\mathbb{E}}}

\theoremstyle{definition}
\newtheorem{definition}{Definition}

\theoremstyle{remark}

\usepackage{tikz}
\usetikzlibrary{fit,calc}

\newcommand*{\tikzmk}[1]{\tikz[remember picture,overlay,] \node (#1) {};\ignorespaces}
\newcommand{\boxit}[1]{\tikz[remember picture,overlay]{\node[yshift=0pt,fill=#1,opacity=.25,fit={($(A)+(0,-0.3\baselineskip)$)($(B)+(.85\linewidth,.1\baselineskip)$)}] {};}\ignorespaces}

\newcommand{\boxitState}[1]{\tikz[remember picture,overlay]{\node[yshift=0pt,fill=#1,opacity=.25,fit={($(A)+(-4.5,-0.5\baselineskip)$)($(B)+(0\linewidth,.1\baselineskip)$)}] {};}\ignorespaces}

\newcommand{\boxitInput}[1]{\tikz[remember picture,overlay]{\node[yshift=0pt,fill=#1,opacity=.25,fit={($(A)+(0,0.4\baselineskip)$)($(B)+(0\linewidth,0\baselineskip)$)}] {};}\ignorespaces}

\newcommand{\boxitStateIn}[1]{\tikz[remember picture,overlay]{\node[yshift=0pt,fill=#1,opacity=.25,fit={($(A)+(0,-0.5\baselineskip)$)($(B)+(0\linewidth,0.1\baselineskip)$)}] {};}\ignorespaces}

\colorlet{myorange}{orange!40}
\colorlet{mygreen}{green!60}
\colorlet{myblue}{blue!30}
\colorlet{mypurple}{purple!30}

\definecolor{blue}{RGB}{0, 0, 255}
\definecolor{orange}{RGB}{255, 165, 0}
\definecolor{green}{RGB}{0, 128, 0}

\title{Best Policy Tracking in Reinforcement Learning}

\author{\name Judith Echevarrieta \email judith.echevarrieta@ehu.eus \\
      \addr University of the Basque Country
      \AND
      \name Etor Arza\email  etor.arza@ntnu.no \\
      \addr Norwegian University of
Science and Technology
      \AND
      \name Aritz Pérez \email aperez@bcamath.org\\
      \addr  Basque Center for Applied Mathematics
      \AND
      \name Josu Ceberio \email josu.ceberio@ehu.eus\\
      \addr  University of the Basque Country\\
      }

\def\month{10}  
\def\year{2026} 
\def\openreview{\url{https://openreview.net/forum?id=b4BbTUF8Xe}} 

\usepackage{fancyhdr}

\fancypagestyle{tmlrheader}{
    \fancyhf{}
    \fancyhead[C]{%
        \small
        Preprint submitted to Transactions on Machine Learning Research (TMLR)\quad\quad
        Accepted article (Author's version)\\
        Reviewed on OpenReview: \url{https://openreview.net/forum?id=b4BbTUF8Xe}
    }
    \renewcommand{\headrulewidth}{0pt}
}

\begin{document}
\pagestyle{tmlrheader}
\thispagestyle{tmlrheader}

\maketitle

\begin{abstract}
Reinforcement learning commonly exhibits non-monotonic performance dynamics in single-task settings. During learning, a sequence of policies is generated whose true performance is unknown and may oscillate significantly across iterations. Despite this well-known phenomenon, standard experimental practice typically reports either the final visited policy or the best among periodically evaluated ones, implicitly assuming that these selection criteria reliably approximate the best-performing policy visited throughout the entire sequence.

In this work, we present an empirical study on the implications of non-monotonicity for policy selection in single-task policy-based reinforcement learning. We introduce post-hoc diagnostic metrics to characterize performance degradation and assess the accuracy and efficiency of different selection criteria. These metrics allow us to evaluate commonly used selection strategies across standard continuous-control benchmarks using a well-established on-policy actor-critic method. Our results show that, in this representative experimental setting, widely adopted selection practices can deviate considerably from the best policy encountered during learning, leading to underperforming estimates and suboptimal use of computational resources. As an alternative, we recommend a validation-based policy selection strategy constrained by computational cost, which provides a more consistent and accurate approximation of the best visited policy while improving selection efficiency. Overall, our findings highlight policy selection as an underexplored factor in the evaluation protocols of reinforcement learning and motivate further study on explicit and standardized validation-based selection procedures.
\end{abstract}

\section{Introduction}\label{SECTION_introduction}
Policy optimization in Reinforcement Learning (RL) consists of finding a decision-making model that allows an agent to maximize a cumulative reward while solving a task through interaction with an environment~\citep{sutton_reinforcement_2018-1}. With the incorporation of neural networks, this learning paradigm has achieved remarkable results in many decision-making and control problems~\citep{zhang_catastrophic_2023,zhang_study_2018}. For instance, methods such as the Deep Q-Network~\citep{mnih_human-level_2015} and Proximal Policy Optimization~\citep{schulman_proximal_2017-1} have shown competitive performance in both discrete and continuous control settings~\citep{bellemare_arcade_2013,mnih_playing_2013-1,arza_performance-guided_2025,makoviychuk_isaac_2021}, while systems such as AlphaZero~\citep{silver_general_2018} have surpassed human-level performance in classic games like chess. These results have led to a primary focus on final performance~\citep{jordan_position_2024}. However, this emphasis may overlook important characteristics of the learning process itself, which may affect the reliability of reported performances and the efficient use of computational resources, both crucial concerns given the high cost of experimentation in RL~\citep{agarwal_deep_2021,jordan_position_2024,patterson_empirical_2024}.

A well-known characteristic of RL is the presence of non-monotonic performance dynamics when solving a specific task~\citep{goos_open_1999,wagner_policy_2014,bertsekas_approximate_2011-1,hasselt_deep_2018}. During each policy optimization process, RL methods generate a sequence of policies as the result of iterative updates driven by experience collected through interaction with the environment. However, in many practical RL settings, the performance of these intermediate policies cannot be directly and exactly evaluated\footnote{Exact policy performance can be computed when the environment dynamics are known and the dimension of the state and action spaces permit exact evaluation. In the remaining cases, policy performance must instead be estimated from the data generated during interactions with the environment, for example through Monte Carlo evaluation~\citep{sutton_reinforcement_2018-1}.}~\citep{zhu_ensuring_2020,sutton_reinforcement_2018-1}, as it depends on an evolving and partially observed learning process~\citep{zhang_catastrophic_2023,liu_measuring_2023,javed_big_2024}. As a result, the true performance of the policies can oscillate over the course of learning~\citep{wagner_policy_2014,zhu_ensuring_2020}, with policies repeatedly improving, deteriorating, and recovering during the optimization process, while updated policies may temporally degrade in performance with respect to the best-performing policy visited so far~\citep{dohare_overcoming_2023}.

Previous research has approached non-monotonicity in RL mainly from three perspectives: understanding its causes~\citep{lyle_disentangling_2024,hasselt_deep_2018,liu_measuring_2023}, mitigating its effects through algorithmic design~\citep{zhang_catastrophic_2023,thomas_data-efficient_2016,schulman_proximal_2017-1,haarnoja_soft_2018}, and developing more rigorous evaluation protocols capable of handling the high variability of RL processes~\citep{islam_reproducibility_2017,khetarpal_re-evaluate_2018,colas_how_2018,patterson_empirical_2024,machado_revisiting_2018,patterson_empirical_2024}. As a result, non-monotonicity is now widely recognized as an intrinsic characteristic of RL learning dynamics rather than an exceptional behavior. However, considerably less attention has been devoted to analyzing its implications for the selection of the policy that is ultimately reported as the result of the learning process.

Under the current state-of-the-art regarding non-monotonicity in RL, a natural question arises: how is the policy ultimately returned as the learned outcome selected from the sequence of visited policies? More importantly, are the selection criteria used in practice able to capture the best performance achieved during learning, while also justifying the invested computational time? If not, these criteria may report a policy with performance below the maximum achieved, thereby attributing to the learning process a lower level of performance than that actually reached. Moreover, this would imply that part of the learning time may have been unnecessary to justify the reported performance, since the same performance could have already been reached at earlier iterations preceding the visit of the best policy.

Despite the possible practical consequences, policy selection is often treated as a secondary implementation detail within experimental settings. The learned policy is commonly chosen either as the final visited model or as the best among periodically evaluated ones~\citep{ouyang_training_2022,le_reasoning_2025,machado_revisiting_2018}, implicitly assuming, without guarantees, that these criteria provide a reliable approximation of the best policy encountered during learning. Moreover, several works have pointed out that, although unstable learning behaviors are clearly present, these phenomena remain insufficiently characterized and measured in RL~\citep{dohare_overcoming_2023,liu_measuring_2023}. At the same time, concerns about reliable selection protocols have increasingly emerged~\citep{chan_measuring_2020}, especially when compared to supervised learning settings where procedures such as cross-validation and separate validation stages are standard practice~\citep{zhang_study_2018}. This work places the overlooked policy selection under non-monotonic learning dynamics at the center of the analysis. Our contributions are as follows:

\begin{enumerate}

\item We explicitly define commonly used policy selection criteria within the standard single-task policy-based RL framework by integrating them into the sequence of policies visited during learning and bringing them from implicit implementation-level heuristics into a unified formal definition.\\[-6mm]

\item We introduce post-hoc diagnostic metrics to characterize performance degradation and to assess the accuracy and efficiency of policy selection criteria in non-monotonic learning processes. These metrics establish a systematic framework for analyzing the reliability of different selection criteria.

\item Based on the proposed metrics, we conduct an empirical analysis of commonly used policy selection criteria in a representative experimental setting, relying on a well-established on-policy actor-critic method and standard continuous-control benchmark environments, both widely used in the RL literature. The considered setting generates diverse policy sequences and degradation patterns, providing broad empirical evidence that characterises the reliability of implicit policy selection practices.\\[-6mm]

\item We identify a systematic gap between the best policy encountered during learning and the policy typically returned by standard selection criteria, showing that policy selection criteria can systematically distort reported performances in RL evaluations.\\[-6mm]

\item Based on the experimental results, we recommend a validation-based selection strategy constrained by computational cost, which provides a more consistent and accurate approximation of the best visited policy while improving selection efficiency. 
\end{enumerate}

Overall, our results show that policy selection under non-monotonic learning dynamics is not a neutral implementation detail, but a factor that can influence both the interpretation of learning performance and the efficiency of experimentation in RL. We hope this work contributes to increasing awareness of how selection decisions may shape reported conclusions, encouraging more explicit and carefully designed selection protocols in future research.

The remainder of this work is organized as follows. Section~\ref{SECTION_RelatedWork} provides an overview of the causes of non-monotonicity in RL, as well as prior work on its measurement and mitigation, together with existing concerns regarding commonly used policy selection criteria in practice. Section~\ref{SECTION_TheoreticalFramework} introduces the background formalization and terminology necessary to define the single-task policy-based framework on which our experimental study focuses, integrating the explicit definition of commonly used policy selection criteria, and introducing the metrics proposed to quantify performance degradation, selection accuracy and selection efficiency. Section~\ref{SECTION_ExperimentalFramework} specifies the design choices used in our experimental analysis, which is presented in Section~\ref{SECTION_ExperimentalAnalysis}, along with the corresponding results and findings. Section~\ref{SECTION_Discussion} discusses the implications of our findings and outlines future research directions. Finally, Section~\ref{SECTION_Conclusions} concludes the paper.

\vspace{-2mm}
\section{Related work}\label{SECTION_RelatedWork}
\vspace{-2mm}

Awareness of non-monotonicity in single-task RL dates back several decades~\citep{goos_open_1999}, where it was recognized as a consequence of the gap between strong theoretical assumptions ensuring monotonic learning guarantees and practical settings that relax these assumptions to increase the applicability of classical RL formulations~\citep{sutton_reinforcement_2018-1,bertsekas_dynamic_2007,dong_simple_2022}.

Modern RL relies on function approximation, data generated by the learning process itself, and bootstrapping~\citep{boyan_generalization_1994,goos_open_1999,hasselt_deep_2018,wissmann_is_2025}, which together introduce several sources of instability. Firstly, the learning objective is estimated from data generated by the current policy, and often even from policies at previous iterations~\citep{thomas_data-efficient_2016}. Furthermore, the data distribution becomes non-stationary and temporally correlated, as it depends on a continuously evolving policy~\citep{zhang_catastrophic_2023,liu_measuring_2023,javed_big_2024,fujimoto_for_2023}. Additionally, stochasticity is intentionally introduced to enable exploration balanced with exploitation~\citep{sutton_reinforcement_2018-1}, further increasing variability during learning. Altogether, the learning process defines a closed-loop system in which the policy generates the data used to update itself~\citep{wagner_policy_2014,han_actor-critic_2020}. All of this breaks assumptions that guarantee monotonic learning, such as data being independent and identically distributed~\citep{zhang_catastrophic_2023,liu_measuring_2023}, while bootstrapping further propagates approximation errors by updating estimates from previous estimates, potentially creating fatal feedback loops~\citep{fujimoto_addressing_2018}. As a result, value-based methods exhibit unstable and divergent learning, with chattering behaviours around a good policy~\citep{goos_open_1999,wagner_policy_2014,bertsekas_approximate_2011-1}. More generally, non-monotonic dynamics can manifest in other forms such as policy collapse, observed in both value-based and policy-based methods, where the learned policy can dramatically worsen after some initial learning as the agent continues to interact with the environment~\citep{dohare_overcoming_2023}.

Several works have attempted to quantify non-monotonicity throughout the learning process of a single task, often using a wide variety of terms to refer to the same underlying phenomenon. Recent studies~\citep{liu_measuring_2023,zhang_catastrophic_2023} measure interference caused by state distribution drift in value-based methods, building on metrics originally developed for catastrophic forgetting in multi-task learning~\citep{serra_overcoming_2018,chaudhry_tiny_2019}. \citet{zhang_catastrophic_2023} argue that temporally correlated and non-stationary training data may lead to catastrophic interference and performance collapse even within single-task learning. They highlight that this can be interpreted as a special case of continual RL, which requires strategies to mitigate interference together with mechanisms to detect and identify sets of states distributed close to each other during single-task learning, playing the role of well-distinguished tasks in multi-task RL~\citep{malagon_self-composing_2024}. \citet{fedus_catastrophic_2020} empirically show how interference arises within Atari games that have long been considered single-task settings. Extending the focus to policy-based methods, \citet{chan_measuring_2020} propose variability and risk metrics for quantifying stability across time during training runs. Despite these efforts in measurement, the effect of non-monotonicity in single-task RL remains unclear~\citep{dohare_overcoming_2023}, with no well-established online characterization or standard measures~\citep{liu_measuring_2023}, especially for policy-based gradient methods~\citep{zhang_catastrophic_2023}.

Another line of work in single-task policy-based RL addresses instability by introducing diverse regularization mechanisms that lead to different algorithmic variants. Examples include constrained policy updates enforcing similarity between consecutive policies via KL divergence~\citep{schulman_trust_2015} or clipping functions~\citep{schulman_proximal_2017-1}, entropy maximization to incentivize more exploratory policies~\citep{haarnoja_soft_2018}, delayed policy updates to mitigate overestimation biases~\citep{fujimoto_addressing_2018}, and replay buffers that store past experiences collected under previous policies to capture the evolving data distribution~\citep{mnih_playing_2013-1,haarnoja_soft_2018,fujimoto_addressing_2018}. Some of these approaches provide theoretical guarantees of monotonic improvement under strong assumptions, such as specific knowledge of environment dynamics~\citep{han_actor-critic_2020}, and accurate value estimation or bounded policy updates~\citep{schulman_trust_2015,eshwar_reliable_2025,zhu_ensuring_2020}. However, in practice these conditions rarely hold~\citep{zhu_ensuring_2020}, resulting in methods that mitigate performance degradation in certain scenarios at the cost of sacrificing other desirable properties, but do not eliminate the non-monotonic dynamics observed during learning~\citep{lyle_disentangling_2024}. For example, \citet{schulman_trust_2015} state that the proposed practical algorithms tend to provide monotonic improvements in simulated robotic tasks, although at the cost of hyperparameter tuning. \citet{dasagi_ctrl-z_2019} argue that while constraining policy updates can successfully prevent large negative performance drops, it can also limit exploration that leads to positive improvements. Overall, these techniques can partially mitigate instability during learning, but leave unaddressed how residual non-monotonicity affects experimental evaluation and policy selection.

At the implementation level, within code-level optimizations~\citep{engstrom_implementation_2020-1}, multiple RL libraries~\citep{raffin_stable-baselines3_2021,petrenko_sample_2020,makoviychuk_isaac_2021} provide heuristic mechanisms for selecting the policy returned as the learned model, often based on periodically saved checkpoints. These approaches either reuse data collected during training or rely on separate evaluation phases to identify the best-performing policy among those visited. However, the reliability and quality of these criteria have rarely been studied systematically~\citep{bellemare_arcade_2013}. Moreover, a common default choice in many implementations is to simply return the final policy~\citep{machado_revisiting_2018,nguyen_uncertainty-guided_2025}, despite the lack of guarantees that it corresponds to the best-performing policy encountered during learning. In addition, learning curves are often constructed using estimates derived from training data to avoid additional evaluation costs~\citep{chan_measuring_2020,machado_revisiting_2018}, even though such estimates have been shown to be optimistic compared to validation-based evaluations~\citep{zhang_study_2018}. These practices on implicit selection criteria in libraries leave open questions about their possible influence on the fair and efficient representation of learning processes under non-monotonic dynamics. \citet{chan_measuring_2020} argue that high instability may imply that RL methods must be checkpointed and evaluated more frequently in order to capture peak performance, which can be expensive. Furthermore, they state that during training, it can be a waste of computational resources to train an unstable method that tends to forget previously learned behaviors.

While prior work has explored different aspects of non-monotonicity in RL, from its underlying causes to partial measurement and mitigation strategies, its impact on the policies selected and reported in practice remains poorly characterized and analyzed. In this work, we present an empirical study focused on measuring the magnitude of non-monotonicity in RL processes, together with a systematic analysis of the evaluation biases and resource inefficiencies introduced by commonly used selection criteria under such dynamics, focusing on standard continuous control benchmarks and a well-established on-policy actor-critic method.

\section{Background and theoretical framework}\label{SECTION_TheoreticalFramework}

In the following, we introduce the formal background and terminology that define the RL framework in which the learning processes analyzed in this work are generated. In addition, we describe the policy selection criteria commonly used in practice to identify the policy returned as the learned model, and define the metrics proposed in this work to quantify non-monotonicity and assess the reliability of these criteria under such phenomenon.
\vspace{-3mm}

\subsection{Interaction-based episodic single-task policy optimization}\label{SECTION_InteractionBasedSTPO}
\vspace{-1mm}

In RL, interaction-based Single-Task Policy Optimization (STPO) consists of finding the explicit decision-making approach that allows an agent to best solve a specific task, based on the experience observed during the interaction with an environment~\citep{sutton_reinforcement_2018-1}. Formally, an environment is defined as a Markov decision process, a tuple $\mathcal{E}=(S,A,p,r)$ where $S$ is the \textit{state space}, $A$ is the \textit{action space}, $p\colon S\times A\rightarrow \Delta(S)$ is the \textit{state transition probability} function that defines a probability distribution $p(\cdot| s,a)$ with support in $S$ for a given state-action pair \mbox{$(s,a)\in S\times A$}, and $r\colon S\times A\rightarrow \mathbb{R}$ is the \textit{reward} function. The agent interacts with the environment according to a probability function \mbox{$\pi\colon S\rightarrow \Delta(A)$}, called the \textit{policy}, which defines a probability distribution $\pi(\cdot|s)$ with support in $A$ for a given state $s\in S$. 

At each step of the interaction, the agent starts from the current state of the environment $s\in~S$ and chooses an action $a\sim\pi(\cdot|s)$. Then, the environment rewards the chosen action $r(s,a)$ and updates the state $s'\sim p(\cdot| s,a)$. The agent-environment interaction breaks down intentionally when completing a number of maximum steps, or naturally upon reaching a terminal state from $S_T\subset S$. In the latter case, the environment is reset before the next interaction, with the initial state drawn from the initial state distribution $p_I\in \Delta(S_I)$, where $S_I\subset S$. The sequence generated by the state-action pairs $\tau=\big((s_0,a_0),...,(s_n,a_n)\big)$ during a $n$ step interaction is named \textit{trajectory}, which constitutes an \textit{episode} when the first state of a trajectory is initial $s_0\in S_I$, the last state-action pair leads to a terminal state $s_{n+1}\in S_T$ and the intermediate states are not terminal $s_i\in S\setminus S_T$ for all $i\in\{1,...,n\}$. The \textit{trajectory return} is given by the discounted cumulative reward of its state-action pairs that, with a slight abuse of notation, we formally denote as
\begin{equation}
r(\tau)= \sum_{i\in\{0,...,\mathsmaller{|}\tau\mathsmaller{|}\}} \gamma^i r(s_i,a_i),
\end{equation}
\vspace{-5mm}

where $\gamma\in[0,1]$ is the \textit{discount factor} and $|\tau|$ is the length of $\tau$, which is a priori unknown for episodes.

In an episodic STPO problem, the objective is to find the policy that maximizes the average episodic return over randomly initialized episodes~\citep{schulman_optimizing_2016}. Since policies are stochastic probabilistic models, in a fixed environment starting from the same initial state, the same policy may generate different episodes. Therefore, the performance of a policy $\pi$ in an environment $\mathcal{E}$ is measured as the expected episodic return, hereafter referred to simply as the \textit{policy return}\footnote{Once the environment is established and known, we omit it from the notation and write $f(\pi):=f(\pi;\mathcal{E})$. For simplicity, and because the experimental analysis is entirely based on this quantity, we subsequently refer to $f(\pi)$ as the \textit{policy return}, or simply \textit{return} when clear from the context.}
\begin{equation}\label{EQUATION_ExpectedDiscountedCumulativeReward}
f(\pi; \mathcal{E})=\expect_{s_0 \sim p_I}~\expect_{\tau \sim (s_0, \pi, p)}~r(\tau),
\end{equation}
\vspace{-5mm}

where $\tau\sim (s_0,\pi,p)$ denotes the episode generated by policy $\pi$ in the environment with state-transition probability distribution $p$, starting from the initial state $s_0\in S_I$. Consequently, the STPO problem is formulated as an optimization problem
\begin{equation}\label{EQUATION_RLproblem}
    \pi^*=\argmax_{\pi\in\Pi}~ f(\pi;\mathcal{E}),
\end{equation}
\vspace{-5mm}

where $\Pi$ is the space of all possible policies.

Interaction-based STPO methods involve learning while interacting with the environment~\citep{sutton_reinforcement_2018-1}. Among them, policy gradient methods stand out, which optimize Equation~\ref{EQUATION_RLproblem} using stochastic gradient descent~\citep{grondman_survey_2012}. We have formalized the optimization process defined by an interaction-based STPO method in Algorithm~\ref{ALGORITHM}. This formulation is intentionally kept abstract to encompass both \textit{on-policy} and \textit{off-policy} learning schemes~\citep{haarnoja_soft_2018,sutton_reinforcement_2018-1}.

\noindent
\begin{minipage}[b]{0.5\textwidth}

\begin{algorithm}[H]
\caption{Interaction-based STPO process}\label{ALGORITHM}
\begin{algorithmic}[1]

\Statex \hspace*{-1.5em} \parbox[t]{1\linewidth}{  \textbf{Input:}\\ 
$\mathcal{E}$: Environment;\\
 $\rho$: Random seed; \\
$t_{max}$: Maximum learning interaction steps;\\[-2mm] }

\Statex \hspace*{-1.5em}  \textbf{Output:} Learned policy 

\vspace{2mm}
\State $t\gets 0$, $D_t\gets\emptyset$
\State $\pi_0\gets$ Initialize policy with $\rho$, $\Pi_t\gets \{\pi_0\}$

\While{$t < t_{max}$}

    \State  $\tau_t\gets$ Interaction of $\pi_t$ with $\mathcal{E}$ during $n$ steps
    \State $D_t\gets D_t\cup\{\tau_t\}$ 

    \State $\widetilde{\pi}_t \gets$ Select best policy in $\Pi_t$ with a criteria

    \State $\pi_t\gets$ Update $\pi_t$ using $D_t$
    \State $t\gets t+ n$
    \State $\Pi_t \gets \Pi_t \cup \{\pi_t\}$

\EndWhile

\State \Return $\widetilde{\pi}_t$

\end{algorithmic}
\end{algorithm}

\end{minipage}
\hfill
\begin{minipage}[b]{0.45\textwidth}
Given an environment $\mathcal{E}$ and starting from a policy $\pi_0$ initialized from a random seed $\rho$ (line~2), the agent–environment interaction and policy update are iteratively repeated until the maximum available interaction steps $t_{max}$ for learning are exhausted (line~3). At each iteration, the gradient of $f(\pi_t)$ is approximated from the trajectory rewards of the finite set of trajectories $D_t$ generated during the previous learning interaction of policies ${\pi_0, ..., \pi_t}$ with the environment $\mathcal{E}$ (lines~4–5), and the policy is updated according to this estimate\footnotemark~(line~7). This results in a sequence of policies $\Pi_t$ that increases with each learning iteration (line~9) and completely determines the \textit{learning process} defined by the method, the environment, and the random seed, to which the policy that is ultimately returned as learned, $\widetilde{\pi}_t$, belongs (lines~6 and 10).
\end{minipage}

\footnotetext{ In on-policy settings~\citep{schulman_trust_2015,schulman_proximal_2017-1}, $D_t$ typically contains data generated by the current policy, whereas off-policy methods~\citep{haarnoja_soft_2018,fujimoto_addressing_2018} may reuse or combine data from previous policies.}

\subsection{Selection criteria}\label{SECTION_SelectionCriteria}

\begin{figure}[H]
\centering
\includegraphics[width=\textwidth]{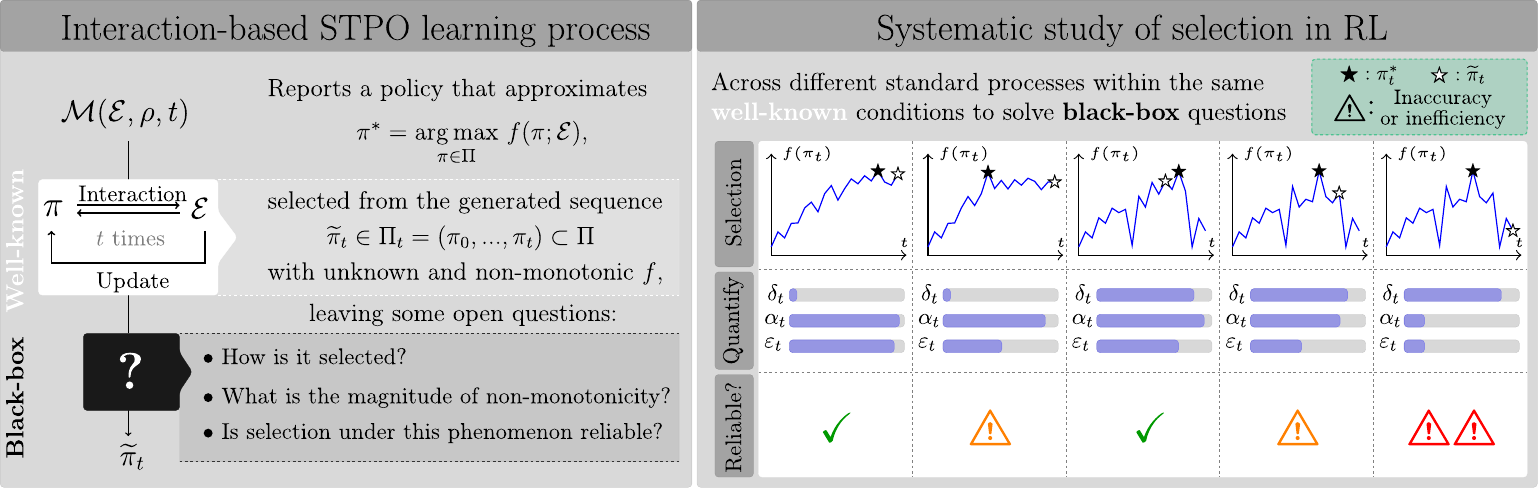}
\caption{The box on the left summarizes the well-known conditions of the learning processes resulting from interaction-based STPO, along with the open questions left by the black-box use of RL libraries in experimental practice. The box on the right illustrates the contribution of our experimental study focused on selection in RL, which analyzes the reliability of commonly used selection criteria by quantifying performance degradation ($\delta_t$), selection accuracy ($\alpha_t$), and selection efficiency ($\varepsilon_t$), across different possible scenarios.}\label{FIGURE_diagram_summary}
\vspace{-5mm}
\end{figure}

In recent years, many libraries~\citep{bou_torchrl_2024,eschmann_rltools_2024,huang_cleanrl_2022,makoviychuk_isaac_2021,petrenko_sample_2020,raffin_stable-baselines3_2021,makoviichuk_rl-games_2021} have been developed, aimed at supporting research in RL. They contain implementations of multiple RL environments and methods, providing baselines to experimentally test the performance of new methods~\citep{chua_deep_2018,pinneri_sample-efficient_2021}. In experimental studies, we observed that these tools are often treated as black boxes. First, an environment $\mathcal{E}$ and a method $\mathcal{M}$ are selected from those implemented by the library. Then, a random seed $\rho$ is set, and the method is run in the environment for a certain time $t$, leading to a learning process. Finally, the policy that is returned by default is used as the learned policy without paying attention to how it was selected from the entire sequence of visited policies. Thus, we denote the anytime learning process as $\mathcal{M}(\mathcal{E},\rho,t)\rightarrow \widetilde{\pi}_t$ (see the box on the left of Figure~\ref{FIGURE_diagram_summary}).

Despite that black-box use of RL libraries, it is well known that during an interaction-based STPO learning process, the exact return of the visited policies is not available due to unknown environment dynamics determined by $p$, which are only partially observable during training~\citep{zhu_ensuring_2020,sutton_reinforcement_2018-1,javed_big_2024}. As described in Algorithm~\ref{ALGORITHM}, policies are updated based on approximations obtained from trajectories collected during finite learning interactions. Furthermore, there is no guarantee that the resulting learning processes will be monotonically increasing~\citep{boyan_generalization_1994}. This means that the evolution of the return across the visited policies $\Pi_t$ does not necessarily improve with each iteration, i.e., there may exist consecutive updates from policy $\pi_i$ to $\pi_j$ such that $f(\pi_i) < f(\pi_j)$. Under these well-known conditions (highlighted in white in Figure~\ref{FIGURE_diagram_summary}, in contrast to black-box practices), the identification of the best policy may not be trivial.

We reviewed some open-source state-of-the-art RL libraries and observed that three simple criteria are implicitly considered for the selection of the output policy. Some of them are based on empirical estimates of the return (see Equation~\ref{EQUATION_ExpectedDiscountedCumulativeReward}) of the policies in the sequence $\Pi_t$, considering the sample mean of the episodic return over a set of episodes $D$ such that
\begin{equation}\label{EQUATION_EmpiricalEstimate}
\widetilde{f}(\pi;D)=\frac{1}{|D|}\sum_{\tau\in D} r(\tau),
\end{equation}
where $\pi\in \Pi_t$.

\textbf{Last visited policy}. Libraries such as Stable Baselines3~\citep{raffin_stable-baselines3_2021} are intentionally designed to continue training based on the current policy state. Therefore, in the absence of an additional policy selection mechanism, they return the last policy in the sequence of visited policies. We call this default selection criterion \textit{last}, which we formally define as
\begin{equation}\label{EQUATION_last}
\widetilde{\pi}_t=\pi_t.
\end{equation}

\textbf{Best policy using training data}. The best policy of the sequence is returned according to the empirical estimate of Equation~\ref{EQUATION_EmpiricalEstimate}, reusing trajectories stored during learning interactions. For example, Sample Factory~\citep{petrenko_sample_2020} and RL Games~\citep{makoviichuk_rl-games_2021} consider a finite number of recent $\kappa$ episodes completed during concatenated trajectories in $D_t$ as the sample $D$. Finally, the selected policy is the one in the sequence with the highest value of that estimate. In this criterion, using a larger number of samples increases the amount of data available for the estimation, at the cost of mixing episodes collected under multiple previously visited policies. We call this criterion \textit{train}, which we formally define as
\begin{equation}\label{EQUATION_train}
\widetilde{\pi}_t=\underset{\pi\in \Pi_t}{\arg\max}~\widetilde{f}(\pi ; D).
\end{equation}

\textbf{Best policy using validation data}. The best policy of the sequence is returned in terms of the empirical estimate of Equation~\ref{EQUATION_EmpiricalEstimate} based on additional validation episodes generated in an independent interaction from training. For instance, Stable Baselines3~\citep{raffin_stable-baselines3_2021} allows evaluating, with a period of interaction steps $\varphi$, the policies using a finite number of additional episodes $\kappa$, which constitute $D$ and are generated by the policies themselves starting from initial states drawn from $p_I$. Finally, the policy with the highest estimate among the evaluated ones is returned. This criterion involves an additional computational overhead proportional to the number of validations and controllable by the evaluation period. We call this criterion \textit{validation}, which we formally define as
\begin{equation}\label{EQUATION_test}
\widetilde{\pi}_t=\underset{\pi_j\in\{\pi_i\in\Pi_t\colon \varphi \mid i\}}{\arg\max}~\widetilde{f}(\pi_j; D).
\end{equation}

In this section, Algorithm~\ref{ALGORITHM} and Equations~\ref{EQUATION_last}, \ref{EQUATION_train}, and \ref{EQUATION_test} formalize the policy selection that is implicitly performed in practice at any time during learning. The equations provide a mathematical definition of the selection criteria, explicitly indicating which policy is ultimately returned from the sequence of visited policies. In practical implementations, however, these criteria rely on equivalent tracking procedures that avoid storing the complete sequence of visited policies. For a detailed description of these practical implementations within the learning process of Algorithm~\ref{ALGORITHM}, we refer the reader to Appendix~\ref{APPENDIX_SelectionCrietriaInPractice}.

\subsection{Measurements for non-monotonicity and selection criteria}
In interaction-based STPO experimental practice, the identification of the best policy is left to the selection criteria described in Section~\ref{SECTION_SelectionCriteria}. However, the magnitude of non-monotonicity remains underexplored~\citep{zhang_catastrophic_2023,liu_measuring_2023}, and the reliability of the policies selected and returned as learned has not been systematically studied, despite previous observations of possible dramatic performance drops~\citep{dohare_overcoming_2023}, optimistic estimates from training data~\citep{zhang_study_2018}, and potentially costly validation procedures~\citep{chan_measuring_2020}. This generates uncertainty in the tracking of the best policy during the learning process. In other words, it is unknown how far the learning curves observed in experimental practice differ from the \textit{ground-truth learning curve}, defined as the curve representing the evolution of $f(\pi^*_t)$ for all iterations\footnote{In Section~\ref{SECTION_InteractionBasedSTPO}, the term $t$ denotes the “number of steps”, so the subindices of the policies that form the sequence $\Pi_t$ are the exact number of steps taken in each iteration. This notation aligns with how interaction-based STPO methods operate in practice, as illustrated in Algorithm~\ref{ALGORITHM}. From Section~\ref{SECTION_ExperimentalFramework} onwards, for the sake of brevity and with a certain abuse of notation, we consider that $t$ represents the number of iterations, and we denote the subindices of the visited policies as $\pi_0,\pi_1,...,\pi_t$, unless explicitly stated otherwise.} $t\leq t_{max}$ such that
\begin{equation}\label{EQUATION_TruthLearningCurve}
\pi^*_t=\underset{\pi\in \Pi_t}{\arg\max}~ f(\pi),
\end{equation}
as well as to what extent these differences may affect the efficiency with which the invested learning time is exploited. To address these issues, we propose three simple post-hoc diagnostic metrics, computed relative to the ground-truth learning curve, to quantify performance degradation and assess the reliability of selection criteria over the sequence of policies generated throughout non-monotonic learning processes (see illustrative examples in the box on the right of Figure~\ref{FIGURE_diagram_summary}).

\begin{definition}
Let $\pi_t$ be the policy visited in the $t$-th iteration of an interaction-based STPO learning process. We define the performance \textit{degradation} of the learning process at iteration $t$ as the overall return decrease relative to the average return achieved so far, which we formally define as
\begin{equation}\label{EQUATION_degradation}
\delta_t=\frac{f(\pi^*_t)-f(\pi_t)}{f(\pi^*_t)-\min\big\{f(\pi_t),\text{mean}_{0,t}\big\}},
\end{equation}
where $\text{mean}_{0,t}:=\operatorname{mean}\{f(\pi_0),\ldots,f(\pi_t)\}$ denotes the mean policy return observed up to time $t$.
\end{definition}

Notice that $\delta_t$ is defined on $[0,1]$ for all $t$, and that at its extreme values, non-monotonicity induces either the largest or smallest decrease in learning performance. When $\delta_t = 1$, at the final learning iteration, the return has decreased to a level equal to or lower than the average return achieved so far. When $\delta_t = 0$, however, the policy obtained in the last iteration achieves the best return observed so far.

\begin{minipage}[b]{0.5\textwidth}

In Figure~\ref{FIGURE_deg}, we illustrate three examples of degradation at iterations 39, 66, and 89. In each case, the light grey vertical band represents the difference between the return of the best policy and the mean return up to that iteration (denominator of Equation~\ref{EQUATION_degradation}), while the darker band within it indicates the difference between the best return and the return of the last policy in the considered iteration (numerator of Equation~\ref{EQUATION_degradation}). At iteration 39, both policies coincide, so degradation is zero. At iterations 66 and 89, the last policy differs from the best one, exhibiting a decrease in return that is respectively greater and smaller than half of the difference with respect to the mean return achieved in each case. Moreover, it can be observed that using the mean return as a reference leads to comparable degradation values for similar return drops across different times, since the mean itself tends to increase throughout learning.
\end{minipage}
\hfill
\begin{minipage}[b]{0.45\textwidth}

\begin{figure}[H]
\centering
\includegraphics[width=\textwidth]{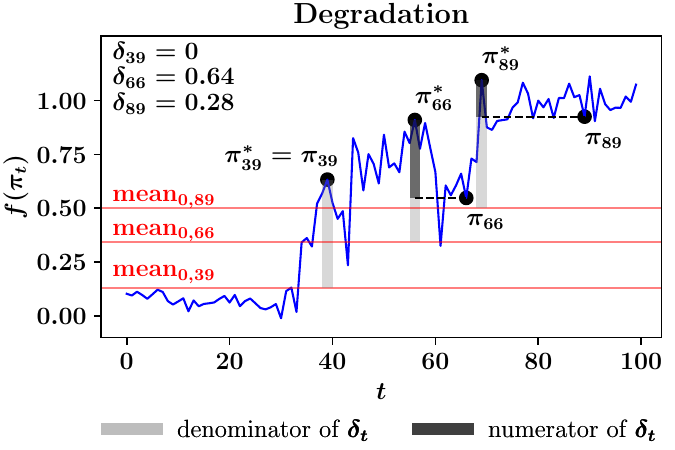}
\caption{Examples of degradation $\delta_t$ (see Equation~\ref{EQUATION_degradation}) at different iterations, over 100 synthetic iterations of policy return.}\label{FIGURE_deg}
\end{figure}
\end{minipage}

Additionally, in order to systematically analyze the quality of the selection, we propose two metrics which, like degradation, depend on the learning iteration and are defined on the interval $[0,1]$.

\begin{definition}
Let $\widetilde{\pi}_t$ be the policy selected and reported as learned after an interaction-based STPO learning process of $t$ iterations. Firstly, we define the \textit{selection accuracy} of the criterion, in terms of performance, as the return of the policy selected by the criterion relative to that of the best policy visited so far, i.e., 
\begin{equation}\label{EQUATION_accuracy}
\alpha_t=\frac{f(\widetilde{\pi}_t)-\min_{0,t}}{f(\pi^*_{t+t'})-\min_{0,t}},
\end{equation}
where $\min_{0,t}:=\min\{f(\pi_0),\ldots,f(\pi_t)\}$ denotes the minimum policy return observed\footnote{The lower end of the normalization is defined using the minimum observed policy return because the absolute minimum achievable return is generally unknown and may differ substantially across environments~\citep{chan_measuring_2020}.} during the learning process up to time $t$ and $t'$ is the additional time spent validating, which only applies to the \textit{validation} criterion. Secondly, we define the \textit{selection efficiency} of the criterion, in terms of time, as the ratio of the total number of learning iterations required to visit, for the first time, a policy with the same or higher return than that of the policy selected by the criterion, i.e., 
\begin{equation}\label{EQUATION_efficiency}
\varepsilon_t=\frac{\min\big\{i\leq t\colon f(\pi_i)\geq f(\widetilde{\pi}_t)\big\}}{t}.
\end{equation}

\end{definition}

\begin{minipage}[b]{0.5\textwidth}

In Figure~\ref{FIGURE_acc}, we illustrate two selections at iteration 65, together with its accuracies. Specifically, $\widetilde{\pi}_{65}$ simulates the policy selected by both the \textit{train} criterion (in orange) and the \textit{validation} criterion (in green). In the first case, $\widetilde{\pi}_{65}$ is compared against the best policy up to iteration $65$, $\pi_{65}^*$. Notice that, the difference between the selected policy and the minimum return up to the current iteration (numerator of Equation~\ref{EQUATION_accuracy}) is less than half of the difference between the best policy and this minimum (denominator of Equation~\ref{EQUATION_accuracy}), meaning that the selected policy recovers less than half of the achievable return. In the second case, $\widetilde{\pi}_{65}$ is compared against the best policy that could be obtained if additional validation time equivalent to $t' = 35$ iterations were used for learning, $\pi^*_{100}$. For this selection, accuracy increases, reaching approximately 76\% of the best achievable return relative to the minimum.
\end{minipage}
\hfill
\begin{minipage}[b]{0.45\textwidth}

\begin{figure}[H]
\centering
\includegraphics[width=\textwidth]{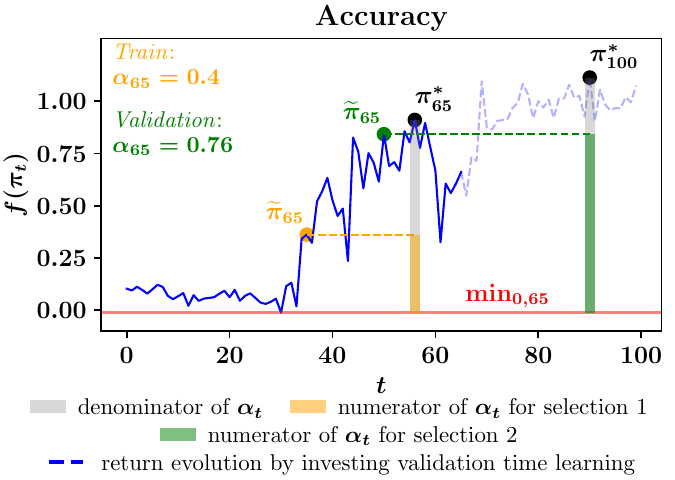}
\caption{Examples of selection accuracy $\alpha_t$ (see Equation~\ref{EQUATION_accuracy}) for two different selections, over 100 synthetic iterations of policy return.}\label{FIGURE_acc}
\end{figure}
\end{minipage}

\begin{minipage}[b]{0.5\textwidth}

Regarding the efficiency measure, in Figure~\ref{FIGURE_eff} we illustrate two examples at iteration 100. Each $\widetilde{\pi}_{100}$ corresponds to a different policy selection under an arbitrary criterion. The green case shows an accurate but less efficient selection, since the same return level was first reached after using only 70\% of the already spent learning iterations, while during the remaining 30\% the return oscillates around similar values due to convergence. The orange case, instead, shows a substantial reduction in efficiency due to selection inaccuracy, as the criterion selects a policy whose return corresponds to early-stage iterations, despite the fact that higher-return policies were visited during the entire 100 iterations of learning.

\end{minipage}
\hfill
\begin{minipage}[b]{0.45\textwidth}

\begin{figure}[H]
\centering
\includegraphics[width=\textwidth]{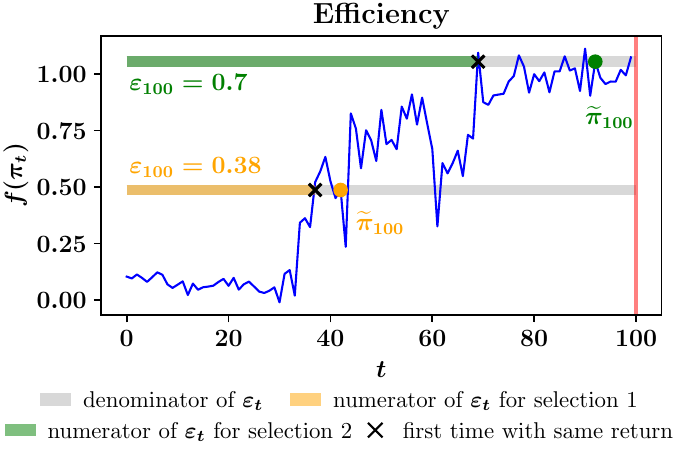}
\vspace{-7mm}
\caption{Examples of selection efficiency $\varepsilon_t$ (see Equation~\ref{EQUATION_efficiency}) for two different selections, over 100 synthetic iterations of policy return.}
\label{FIGURE_eff}
\end{figure}
\vspace{-5mm}
\end{minipage}

As the green example in Figure~\ref{FIGURE_eff} already suggests, an accurate selection does not necessarily imply an efficient one. The proposed accuracy and efficiency measures therefore capture different aspects of the reliability of policy selection. Consequently, both metrics must be considered jointly to properly assess the quality of a selection criterion.

\section{Experimental framework}\label{SECTION_ExperimentalFramework}

In this section, we describe the experimental setup used within the interaction-based STPO framework to carry out the analysis of selection in RL.

\subsection{Learning processes from standard setups}
In this work, we analyze learning processes within standard interaction-based STPO setups. We focus on Proximal Policy Optimization~\citep{schulman_proximal_2017-1} method, which is one of the on-policy actor-critic reference methods~\citep{arza_performance-guided_2025,makoviychuk_isaac_2021}, and we consider its open-access implementation in Stable-Baselines3 library~\citep{raffin_stable-baselines3_2021}. The study is conducted on widely used continuous-control benchmark environments, specifically Box2D environments LunarLanderContinuous (LLC) and BipedalWalker (BW), and MuJoCo environments Swimmer (S), HalfCheetah (HC), Hopper (H), Ant (A), and Walker2d (W2d), all of which are available in the Gymnasium library~\citep{towers_gymnasium_2025}. To ensure comparability with experimental practice and avoid introducing biases in the magnitude of degradation, we rely on the default configurations provided by these libraries, complemented with hyperparameters tuned in RL Baselines3 Zoo~\citep{raffin_rl_2020}. Following current recommendations for robust RL experimentation~\citep{jordan_position_2024,patterson_empirical_2024}, we intentionally minimize intervention in the configuration of the learning processes, and the few modifications considered are detailed in Appendix~\ref{APPENDIX_SetupConfiguration}.

\subsection{Ground truth}

To quantify degradation $\delta_t$, accuracy $\alpha_t$, and efficiency $\varepsilon_t$ throughout the learning processes, we require an estimate of the return (see Equation~\ref{EQUATION_ExpectedDiscountedCumulativeReward}) of the visited policies $\Pi_t$. Since the exact value of $f(\pi)$ for each $\pi\in \Pi_t$ cannot be computed in practice due to the unknown environment dynamics and the potentially infinite number of episodes generable by each policy, we approximate it using the empirical estimate defined in Equation~\ref{EQUATION_EmpiricalEstimate}. For each environment, we fix a large subset of 500 initial states and estimate the return of each policy from the episodes generated by the policy starting from those states.

The sample mean, when used as a ground truth estimate, remains a point estimate and does not explicitly capture the uncertainty of the underlying random variable~\citep{agarwal_deep_2021}. Nevertheless, since episodes are generated by the same policy $\pi$ under a fixed initial state distribution $p_I$, the sample mean constitutes an unbiased estimator. Furthermore, by considering a large number of episodes, the variance of the estimator is small compared to the estimates typically used by the \textit{validation} criterion. At the same time, we use the same initial states for all policies in the sequence $\Pi_t$, which leads to paired comparisons of returns between policies, with lower variance in performance differences across learning iterations. Finally, in order to ensure a fair comparison between ground truth and validation~\citep{zhang_study_2018}, notice that the subset of initial states considered for the ground truth estimate differs from the one used during validation in the \textit{validation} criterion. Additional details and empirical evidence supporting the stability of this estimate are provided in Appendix~\ref{APPENDIX_EmpiricalEstimates_GroundTruth}.

\subsection{Representation of ground truth, degradation, accuracy and efficiency}
Given a method, an environment and a runtime, different random seeds can define completely different ground-truth learning curves~\citep{cheng_control_2019,henderson_deep_2018-1}. In fact, the random seed is one of the factors causing variability between learning processes in RL~\citep{chan_measuring_2020}. To account for this variability in our analysis, following best practice guidelines in RL~\citep{agarwal_deep_2021,chan_measuring_2020}, we consider 30 random seeds for each combination of method and environment, and we represent the ground truth $f (\pi_t)$, degradation $\delta_t$, accuracy $\alpha_t$ and efficiency $\varepsilon_t$ using the median and interquartile range across the seeds. 

Moreover, since the defined diagnostic metrics depend on the number of learning iterations $t$, each learning process leads to multiple values for each metric, one per iteration. However, the behaviour of the policy return is not homogeneous throughout the learning process, and both the selection criteria and the proposed metrics are inherently dependent on this behaviour. To facilitate the interpretation and aggregation of the results, we apply a post-hoc partition of each learning process based on the evolution of the policy return, resulting in three regions, namely \textit{initialization}, \textit{learning}, and \textit{convergence}. This partition allows metric values to be grouped according to the different stages of the learning process, providing a more interpretable characterization of the reliability of the selection criteria under different learning dynamics, in line with the recommendations of \citet{chan_measuring_2020}. Specifically, for each region, we report the mean and standard deviation of the values of each time-dependent metric across the iterations within that region.

\begin{figure}[H]\label{FIGURE_regions}
\centering
\includegraphics[width=\textwidth]{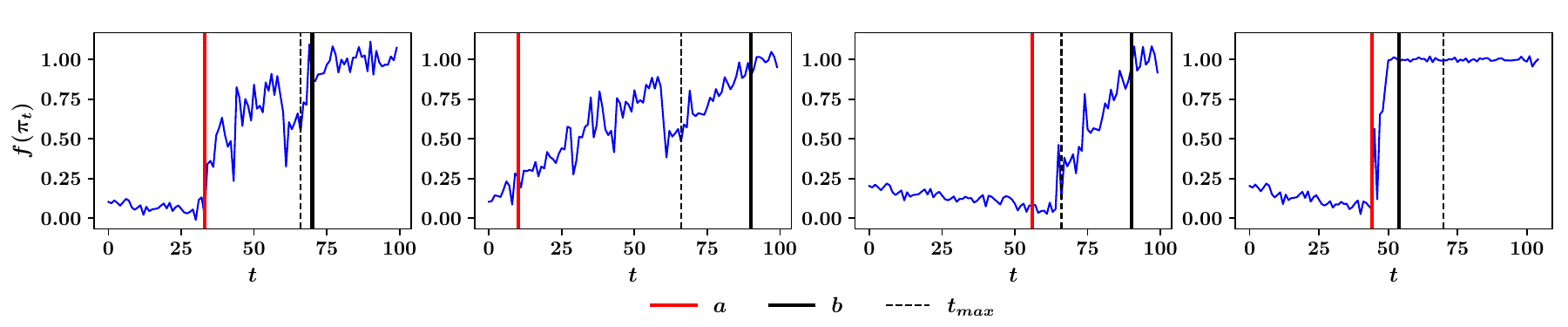}
\vspace{-6mm}
\caption{Characterization of the regions according to Equation~\ref{DEFINITION_regions} across four different cases, with 100 synthetic iterations of the policy return and a window size of $w=10$. From left to right: $a$ and $b$ determined by the statistical condition in $A$ and $B$; $a$ and $b$ imposed by the minimum size condition; $a$ truncated not to exceed $t_{max}$; and $b$ truncated to ensure a well-defined interval $[a,b]$ with a minimum length of 10\%.}\label{FIGURE_regions}
\vspace{-2mm}
\end{figure}

Intuitively, the initialization phase corresponds to the early stage of training, where performance is highly unstable and no consistent improvement is observed. The learning phase captures the period in which the policy progressively improves, although non-monotonic behavior may still occur. Finally, the convergence phase corresponds to the stage where performance stabilizes and no further significant improvements are observed. We delimit the transition between these regions by two iteration indices, the \textit{learning start} $a$ and the \textit{convergence start} $b$, which we formally define as follows

\begin{equation}\label{DEFINITION_regions}
\begin{aligned}
a &= \min\big\{\max\{ t\in A : \exists i\in A,\ t-w\le i< t\}\cup \{ w \}, t_{\max}-w\big\}, \\
b &= \max\big\{\min\{ t\in B : \exists i\in B,\ t< i \le t+ w \}\cup \{T-w\}, a+w\big\},
\end{aligned}
\end{equation}

where $A=\{t\colon f(\pi_t)-\min_{0,t}<\text{std}_{0,t}\}$ contains iterations in which the current return remains within the observed variability (indicating that no clear improvement has yet occurred), and ${B=\{t\colon \max_{t,T}-f(\pi_t)<\text{std}_{t,T}\}}$ contains iterations close to the best future return values (indicating proximity to convergence). Here, $T=1.5\cdot t_{\max}$ denotes the maximum execution time extended by half to favour the apparition of convergence in the generated processes. The shorthand notation $\min_{i,j}$, $\max_{i,j}$, and $\text{std}_{i,j}$ denotes the minimum, maximum, and standard deviation, respectively, over the sequence of policy returns $\{f(\pi_i),\ldots,f(\pi_j)\}$.

Notice that, in Equation~\ref{DEFINITION_regions}, we impose a degree of local persistence on the conditions defining $A$ and $B$ over a window of length $w=\lfloor 0.1\cdot T\rfloor$, corresponding to 10\% of the learning horizon, since these conditions may not hold over consecutive iterations in practice. In this work, $w$ is chosen to ensure that the resulting regions match their intended interpretation across the generated sequences included in the analysis, as illustrated in Appendix~\ref{APPENDIX_regions}. Furthermore, the unions guarantee a minimum size for the regions, while the minimum and maximum operations in the definitions of $a$ and $b$, respectively, ensure that $a<t_{\max}$ and $b>a$. These properties support the intended characterization of the regions $[0,a]$, $[a,b]$, and $[b,T]$ as initialization, learning, and convergence, respectively. See Figure~\ref{FIGURE_regions} and Appendix~\ref{APPENDIX_regions} for synthetic and benchmark examples, respectively. To preserve this interpretation, the value of $w$ may need to be adjusted when applying this partition to different policy sequences.

\subsection{Configuration of the selection criteria}\label{SUBSECTION_CrietriaConfiguration}

The \textit{last} criterion is the simplest of the three described in Section~\ref{SECTION_SelectionCriteria}, and does not require the definition of any parameters for its application. However, the \textit{train} criterion depends on the number of most recent episodes $\kappa$ completed during learning for the estimation $\widetilde{f}$, and in the case of \textit{validation}, in addition to the number of validation episodes $\kappa$, the validation period $\varphi$ must also be determined.

To carry out the experimental analysis, we set the number of episodes $\kappa$ for the \textit{train} criterion by evaluating the selection accuracy (see Equation~\ref{EQUATION_accuracy}) across the three regions of the different environments on a grid of possible values, ultimately selecting the value $\kappa=50$ that yields the highest average accuracy. For the \textit{validation} criterion, the parameter $\kappa$ determines the number of validation episodes to be considered in an additional interaction that involves an extra computational cost. Validation with the same number of episodes from different policies does not necessarily imply the same cost, so the same pairs of $\kappa$ and $\varphi$ may lead to different validation costs in different processes. To promote a controlled and comparable cost across processes, for a fixed $\kappa$ we determine the validation period automatically based on a predetermined threshold $\varphi_c$ for the ratio between validation cost and learning cost. In this way, the resulting \textit{validation} criterion with fixed parameters $\kappa$ and $\varphi_c$ adapts the validation period dynamically, to keep the validation-to-learning cost ratio close to $\varphi_c$ throughout the learning process. We select the pair $\kappa=5$ and $\varphi_c=0.2$ from a grid with which we observe the highest average accuracy. 

The exact practical implementation considered for the \textit{validation} criterion is described in Appendix~\ref{APPENDIX_SelectionCrietriaInPractice}, while Appendix~\ref{APPENDIX_CriteriaConfiguration} details the selected configurations, discusses the associated validation costs and reports a sensitivity analysis across multiple validation-to-learning cost thresholds.

\section{Experimental analysis}\label{SECTION_ExperimentalAnalysis}

Building on the established experimental framework, this section presents a sequence of complementary analyses addressing the same central question: whether the policy selection criteria implicitly used in experimental practice provide reliable policy selections under non-monotonic learning, and how their reliability relates to degradation. First, we characterize the degradation exhibited by the learning processes considered in the study, establishing the empirical conditions under which policy selection takes place. Second, we assess the commonly used selection criteria in terms of selection accuracy and efficiency, characterizing how reliably they identify best visited policies and how this selection reflects the computational effort invested during learning. Third, we analyze the relationship between degradation and the accuracy and efficiency of the selection criteria to determine whether their reliability changes systematically with the severity of degradation. Finally, we compare the selection criteria conditioned on the observed degradation levels to characterize which criteria are more appropriate under different degradation conditions. 

The experimental analysis is supported by figures that are first described in terms of what they show and subsequently interpreted to extract the main findings. Detailed descriptions of the construction underlying each figure are provided in Appendix~\ref{APPENDIX_FigureConstruction}.

\subsection{Distribution of degradation, accuracy and efficiency}\label{SECTION_Distribution_deg_acc_eff}
\begin{figure}[H]
\centering
\includegraphics[width=\textwidth]{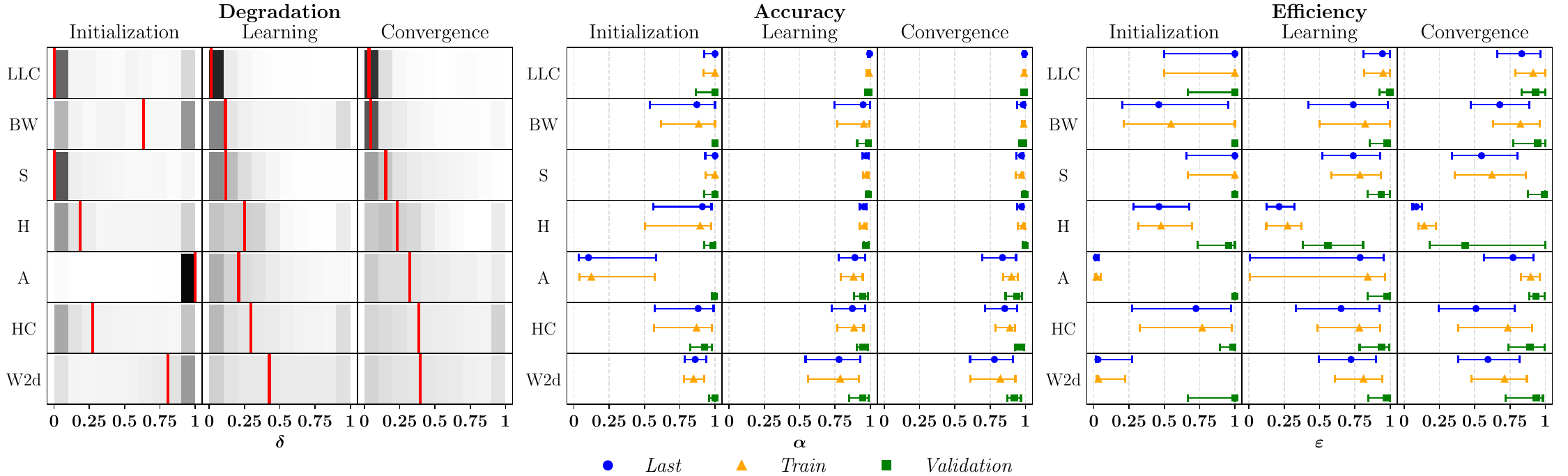}
\caption{On the left, distributions of degradation in each region for each environment shown as a color-coded histogram with medians in red. In the centre and on the right, distributions of accuracy and efficiency respectively for each selection criterion, with the interquartile range and the median. See Appendix~\ref{APPENDIX_FigureConstruction_Complete} for details on figure construction.}\label{FIGURE_analysis_deg_acc_eff}
\vspace{-5mm}
\end{figure}

The first plot of Figure~\ref{FIGURE_analysis_deg_acc_eff} shows the distribution of degradation (see Equation~\ref{EQUATION_degradation}) for each environment and region across the 30 generated policy sequences. Each distribution is represented by a histogram with 10 bins, where darker colors indicate a larger number of degradation samples within the corresponding interval, and the median is highlighted in red. In general, across all environments and regions, median degradation values are below 0.5 in 18 out of 21 cases, indicating that high degradation levels are relatively uncommon across the considered scenarios.

We order environments by their median degradation across the three regions to facilitate the analysis of its influence on selection accuracy and efficiency (see Table~\ref{TABLE_Env_Complexity}). Interestingly, we observe that LLC and BW exhibit the lowest degradation levels, whereas A, HC, and W2d tend to present higher values.

\begin{minipage}[b]{0.5\textwidth}
 This ordering is broadly consistent with the relative physics simulation complexity of the environments. As described in their original formulations by~\citet{towers_gymnasium_2025}, the former are based on the Box2D engine, which involves simpler toy-game-based 2D rigid-body simulations for control tasks, whereas the latter belong to the MuJoCo engine, which represents more challenging robotics-based problems with higher dimensionality and non-linear dynamics. We note that environments with lower degradation are typically those described as simpler in terms of simulation engine and state-action dimensionality, while higher degradation values tend to appear in environments considered more complex according to standard benchmark descriptions.

\end{minipage}
\hfill
\begin{minipage}[b]{0.45\textwidth}

\begin{table}[H]
\caption{Environments ordered according to the median degradation across 30 policy sequences, together with their characteristics related to the physics simulation complexity. This ordering is used consistently across all figures in the analysis.}\label{TABLE_Env_Complexity}

\centering
{\footnotesize
\begin{tabular}{|c|c|c|c|c|c|}
\hline 
$\mathcal{E}$ & Engine & $|S|$ & $|A|$ & Median $\delta$ \\ 
\hline 
LLC & Box2D & 8 & 2 & 0.02 \\ 
\hline 
BW & Box2D & 24 & 4 & 0.10 \\ 
\hline 
S & MuJoCo & 8 & 2 & 0.12 \\ 
\hline 
H & MuJoCo & 11 & 3 & 0.23\\ 
\hline 
A & MuJoCo & 27 & 8 &  0.30\\ 
\hline 
HC & MuJoCo & 17 & 6 & 0.32\\ 
\hline 
W2d & MuJoCo & 17 & 6 &  0.44\\ 

\hline 
\end{tabular} 
}
\end{table}
\vspace{1mm}

\end{minipage}

The most pronounced differences of degradation between environments occur during initialization. In this region, degradation values are closer to 1, particularly for BW, W2d, and A. These environments also correspond to those with the highest overall degradation levels within each engine group (see Table~\ref{TABLE_Env_Complexity}), suggesting that more complex environments require a longer initialization phase before the learning process begins to consistently improve upon the initial random policy. Simpler environments, such as LLC and S, instead tend to exhibit more stable behaviour under random policies, leading to lower degradation during the earliest stages of learning. This interpretation is further supported in Table~\ref{TABLE_regions} of Appendix~\ref{APPENDIX_regions}, where larger learning start limits can be observed for the environments with higher degradation. 

\vspace{-1mm}
\begin{minipage}[b]{0.5\textwidth}
The remaining plots of Figure~\ref{FIGURE_analysis_deg_acc_eff} show the selection accuracy (see Equation~\ref{EQUATION_accuracy}) and efficiency (see Equation~\ref{EQUATION_efficiency}) of the three criteria for each environment and region. In these cases, distributions are reported as segments, where the central marker indicates the median and the interval represents the interquartile range. Table~\ref{TABLE_deg_acc_eff} complements and summarizes these results by providing a compact quantitative overview of the corresponding distributions, dividing them into three ranges delimited by the values 0.25 and 0.75.

\end{minipage}
\hfill
\begin{minipage}[b]{0.45\textwidth}
\begin{table}[H]
\centering
\caption{Count of environment-region cases where median accuracy ($\alpha$) or efficiency ($\varepsilon$) over 30 policy sequences falls within the low $[0,0.25)$, middle $[0.25,0.75)$, or high $[0.75,1]$ ranges.}\label{TABLE_deg_acc_eff}
{\footnotesize
\begin{tabular}{|c|c|c|c|c|c|}
\hline
Metric & Criterion & Low & Middle & High \\
\hline
\multirow{3}{*}{$\alpha$} & \textit{{\color{blue}last}} & 1 & 0 & 20  \\
 & \textit{{\color{orange}train}} & 1 & 0 & 20 \\
 & \textit{{\color{green}validation}} & 0 & 0 & 21  \\

  \hline
\multirow{3}{*}{$\varepsilon$} & \textit{{\color{blue}last}} & 4 & 11 & 6  \\
 & \textit{{\color{orange}train}} & 3 & 6 & 12 \\
 & \textit{{\color{green}validation}} & 0 & 2 & 19 \\

\hline
\end{tabular}

}
\end{table}

\end{minipage}

In terms of accuracy, most criteria achieve values above 0.75 in 61 out of 63 cases. The \textit{validation} criterion achieves the highest values above this threshold in all cases (21/21). In contrast, \textit{last} and \textit{train} exhibit lower median accuracies. This trend is more pronounced during initialization, particularly in the case of A, with a median lower than 0.25, which coincides with the highest degradation levels. A similar pattern is observed for efficiency, although with lower values overall (compare the rows of the column named \textit{High} of Table~\ref{TABLE_deg_acc_eff}). Efficiency is above 0.75 in 36 out of 63 cases. The \textit{validation} criterion again shows consistently high values (19/21), whereas \textit{last} and \textit{train} fall below 0.75 in a larger number of cases (15/21 and 9/21, respectively). In contrast to accuracy, the few cases below 0.25 include environment H, which corresponds to the lowest convergence start limits among all environments (see Table~\ref{TABLE_regions} in Appendix~\ref{APPENDIX_regions}), explaining the decrease in efficiency due to an excessively long learning horizon.

\subsection{Relationships between degradation and selection reliability}

\begin{figure}[H]
\vspace{-2mm}
\centering
\includegraphics[width=\textwidth]{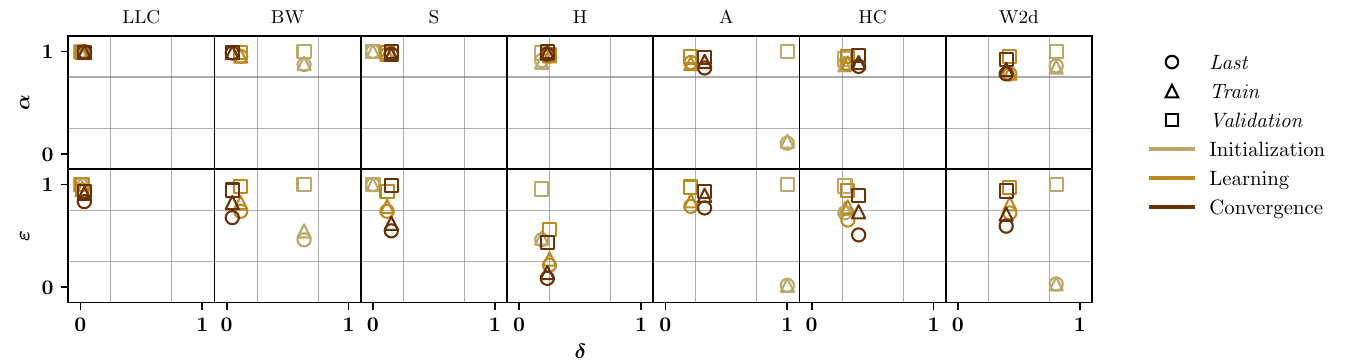}
\vspace{-5mm}
\caption{Scatter plots of median degradation, $\delta$, versus selection accuracy, $\alpha$, (first row) and efficiency, $\varepsilon$, (second row) for the different selection criteria (distinct markers), across regions (colors) and environments (columns). See Appendix~\ref{APPENDIX_FigureConstruction_Complete} for details on figure construction.}\label{FIGURE_ScatterPlots_deg_acc_eff}
\vspace{-6mm}
\end{figure}

To analyze the relationship between degradation and the reliability of the selection criteria, Figure~\ref{FIGURE_ScatterPlots_deg_acc_eff} shows scatter plots of accuracy and efficiency against degradation, considering all regions for each environment. Each point corresponds to a criterion-region pair within an environment, and its coordinates are determined by the median values of the corresponding metrics. A grid is included to mark the three value ranges (low, middle and high), delimited by the thresholds 0.25 and 0.75, to facilitate the analysis.

The scatter plots make explicit several of the patterns already suggested in Section~\ref{SECTION_Distribution_deg_acc_eff}. In the first row, accuracy remains generally high, with lower values mainly corresponding to the \textit{last} and \textit{train} criteria in initialization regions with higher degradation levels. In contrast, the \textit{validation} criterion remains consistently accurate across all environments, regions, and degradation levels. This suggests that the selection capability of the \textit{last} and \textit{train} criteria deteriorates in scenarios with higher degradation, where identifying the best policy becomes more challenging. Instead, in near-zero degradation cases such as LLC, more sophisticated criteria like \textit{train} and \textit{validation} are not necessary, as the simpler \textit{last} criterion already achieves accurate selections.

In the second row, while \textit{validation} maintains high efficiency levels, both \textit{last} and \textit{train} show a decrease even under moderate degradation levels. Notably, this drop in efficiency is not solely linked to selection inaccuracies, as accuracy remains comparatively high in many of these cases. There is a second source of inefficiency mainly associated with the use of excessively long learning horizons, particularly in environments with sooner convergence starts, such as H. The two sources of inefficiency coincide with those illustrated in the examples of Figure~\ref{FIGURE_eff}.

An additional observation is the strong similarity between \textit{last} and \textit{train} across both accuracy and efficiency. This is consistent with the fact that \textit{train} is based on empirical estimates of the policy return computed from trajectories generated during training, which are the same data used for policy updates in on-policy RL methods. This shared data source may limit the ability of the \textit{train} criterion to track occasional performance peaks during the learning process, with estimates that may inherit part of the bias introduced during policy updates. This phenomenon is related to overfitting effects in RL~\citep{zhang_study_2018}, where return estimates based on training data fail to generalize to the true underlying policy return. A more detailed analysis of training-based estimations is provided in Appendices~\ref{APPENDIX_TrainTestEstimates} and~\ref{APPENDIX_TrainOverfitting}.

\subsection{Comparative analysis of criteria}
\vspace{-7mm}

\begin{minipage}[b]{0.5\textwidth}
We continue the analysis by performing pairwise comparisons between the selection criteria. Figure~\ref{FIGURE_comparison_times} shows the proportion of times each criterion selects a policy with higher return than its counterpart.
\vspace{2mm}

Consistent with the previous analysis, the \textit{validation} criterion achieves the highest proportion of wins. Averaged across all environments and regions, it outperforms both \textit{last} and \textit{train} in approximately 66\% of the comparisons. This trend is also reflected in Figure~\ref{FIGURE_comparison_times}, where the green segments corresponding to \textit{validation} are predominant. Furthermore, the proportion of ties, represented by the white spaces, averages only 8\% in comparisons involving \textit{validation}, compared to 31\% in the \textit{last} versus \textit{train} comparison. This indicates that \textit{validation} not only wins more frequently, but also tends to do so by a larger margin than the other criteria.
\end{minipage}
\hfill
\begin{minipage}[b]{0.45\textwidth}

\begin{figure}[H]
\centering
\includegraphics[width=\textwidth]{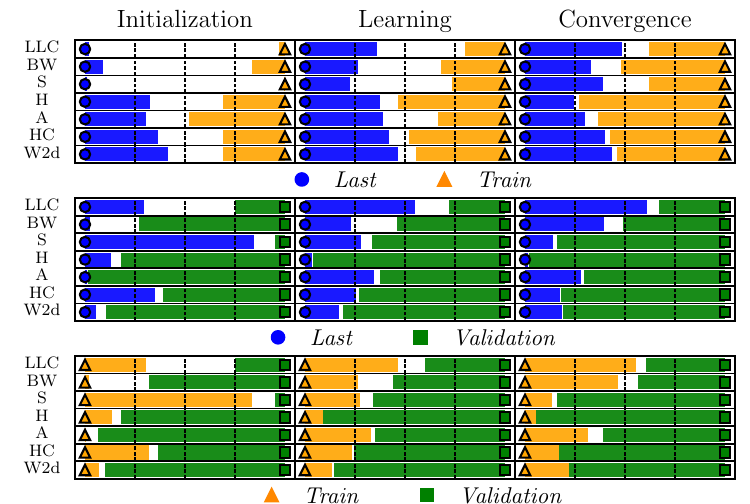}
\caption{Proportion of times each criterion outperforms the other in pairwise comparisons by region, with ties in white. Dashed vertical lines highlight the proportions 0.25, 0.5, and 0.75. See Appendix~\ref{APPENDIX_FigureConstruction_Pairwise} for details on figure construction.}
\label{FIGURE_comparison_times}
\end{figure}
\end{minipage}

\begin{figure}[H]
\centering
\includegraphics[width=\textwidth]{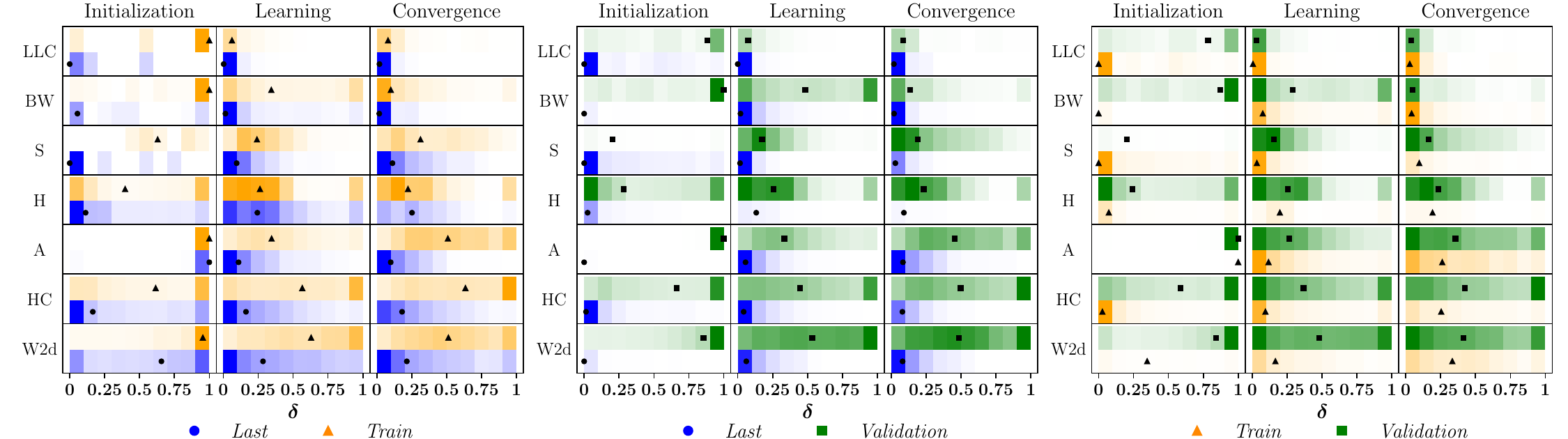}
\caption{Pairwise comparison of selection criteria across degradation levels for different environments and regions. For each environment-region pair, rows correspond to compared criterion pairs and columns to degradation intervals. Color intensity encodes both win proportions (within each cell) and the number of comparisons (across cells) via joint normalization, while markers indicate median degradation values for each criterion. See Appendix~\ref{APPENDIX_FigureConstruction_Pairwise} for details on figure construction.}\label{FIGURE_comparison_deg}
\vspace{-5mm}
\end{figure}

To further analyze how degradation affects the relative performance of the selection criteria, Figure~\ref{FIGURE_comparison_deg} shows the outcomes of the pairwise comparisons conditioned on degradation levels. For each environment-region pair, each row corresponds to a pair of criteria distinguished by their corresponding colors, while columns represent degradation intervals. Within each cell, the color intensity indicates the proportion of cases in which one criterion outperforms the other. For each pair of criteria, we count the number of cases in which either criterion outperforms the other within each degradation interval (excluding ties). Color intensities are then normalized jointly across both criteria and all degradation intervals using a common maximum frequency.  As a result, the figure simultaneously reflects two aspects: the relative proportion of wins between the two criteria within each degradation interval, and the relative number of observed comparisons across degradation levels. Consequently, darker colors have two interpretations. Within each cell, they indicate a higher probability of one criterion outperforming the other. Across cells, they also indicate that a larger number of comparisons fall within the corresponding degradation range. Table~\ref{TABLE_ComparisonByDegradationLevel} provides the corresponding numerical values by reporting the proportion of pairwise comparisons won by each criterion across three degradation ranges delimited by 0.25 and 0.75 values and also by the three learning regions.

Several clear patterns can be observed. First, the advantage of \textit{validation} over both \textit{last} and \textit{train} increases with degradation. Considering all regions together, \textit{validation} outperforms \textit{last} in 66\% of low-degradation cases, 95\% of middle-degradation cases, and all high-degradation cases (100\%). A similar trend is observed against \textit{train}, where the winning proportion increases from 75\% to 88\% and 92\% across the same degradation ranges. These results indicate that the benefits of validation-based selection become increasingly important as degradation grows. In contrast, the differences between \textit{train} and \textit{last} are considerably smaller. Across all regions, \textit{train} outperforms \textit{last} in 44\% of low-degradation cases, 61\% of middle-degradation cases, and 80\% of high-degradation cases. Although \textit{train} tends to perform better on average, the gap remains substantially smaller than that observed for comparisons involving \textit{validation}, reinforcing the similarity between both criteria already observed in Figures~\ref{FIGURE_analysis_deg_acc_eff} and~\ref{FIGURE_ScatterPlots_deg_acc_eff}.

\begin{table}[H]
\centering
\caption{Proportion of pairwise wins between selection criteria across learning regions (initialization, learning and convergence) and degradation ranges (low, middle and high).}\label{TABLE_ComparisonByDegradationLevel}
{\footnotesize
\begin{tabular}{|c|ccc|ccc|ccc|ccc|}
\hline
\multirow{2}{*}{Case} & \multicolumn{3}{c|}{Initialization} & \multicolumn{3}{c|}{Learning} & \multicolumn{3}{c|}{Convergence} & \multicolumn{3}{c|}{All regions} \\
 & Low & Middle & High & Low & Middle & High & Low & Middle & High & Low & Middle & High \\
\hline
\textit{{\color{orange}train}} wins \textit{{\color{blue}last}} & 0.33 & 0.41 & 0.62 & 0.33 & 0.57 & 0.80 & 0.58 & 0.68 & 0.90 & 0.44 & 0.61 & 0.80 \\
\textit{{\color{green}validation}} wins \textit{{\color{blue}last}} & 0.55 & 0.89 & 0.99 & 0.56 & 0.94 & 1.00 & 0.81 & 0.97 & 1.00 & 0.66 & 0.95 & 1.00 \\
\textit{{\color{green}validation}} wins \textit{{\color{orange}train}} & 0.62 & 0.86 & 0.96 & 0.69 & 0.88 & 0.91 & 0.86 & 0.88 & 0.91 & 0.75 & 0.88 & 0.92 \\
\hline
\end{tabular}
}
\end{table}

Overall, the results highlight the importance of validation-based selection criteria, which are considerably more robust to degradation than those relying exclusively on training trajectories or on the final visited policy in the learning process.

\section{Discussion and future directions}\label{SECTION_Discussion}

The analysis presented in this work focuses on the best policy selection criteria in interaction-based episodic single-task policy optimization. By explicitly characterizing the degradation patterns along policy sequences and evaluating how different selection criteria respond to them, we provide a framework for understanding the conditions under which best-policy tracking becomes reliable. Beyond the experimental findings themselves, several broader implications and future research directions emerge from this analysis.

\subsection{Best-policy tracking beyond on-policy benchmark settings}

The empirical study presented in this work focuses on Proximal Policy Optimization~\citep{schulman_proximal_2017-1} and a collection of standard continuous-control benchmark environments from Box2D and MuJoCo~\citep{towers_gymnasium_2025}. These choices provide a representative experimental setting in which the policy selection problem can be studied under a wide range of degradation patterns. In particular, the considered algorithm-environment combinations generate a diverse collection of policy sequences, ranging from nearly monotonic to highly degraded learning processes across the initialization, learning, and convergence regions. This diversity provides a broad empirical basis for analyzing the reliability of the commonly used selection criteria under different degradation conditions.

In contrast, the proposed framework is not specific to the algorithm or environment used to generate the policy sequences. Different combinations of learning algorithms and environments can act as different generators of policy sequences and may therefore produce different degradation patterns. The conclusions of the analysis, however, are conditioned on the degradation exhibited by the generated policy sequence rather than on the mechanism used to generate it. In particular, when degradation is negligible, the simpler \textit{last} criterion may provide accurate selections without requiring the additional computational cost associated with validation-based selection, whereas the benefit of the \textit{validation} criterion becomes increasingly relevant as degradation increases. Thus, while the empirical evidence presented in this work is focused on a representative set of learning processes, the conclusions observed can provide a reference for other learning settings exhibiting comparable degradation levels. 

The generality of the proposed framework provides a natural basis for building on the presented analysis, including off-policy methods~\citep{haarnoja_soft_2018,fujimoto_addressing_2018}, discrete-action environments~\citep{mnih_playing_2013-1,bellemare_arcade_2013}, and more realistic domains~\citep{makoviychuk_isaac_2021}. For instance, off-policy algorithms can reuse previously collected experience through replay buffers, allowing multiple learning updates to be performed from previously collected interactions. Consequently, for a given interaction budget, they may generate longer sequences of visited policies than the on-policy processes considered in our experiments. This difference can affect the configuration of the \textit{validation} criterion, since, for longer policy sequences, the validation period could be increased to preserve the specified validation-to-learning cost ratio, such as the 20\% threshold used in our experiments, while potentially affecting selection accuracy. Thus, extending the study to policy sequences generated by other methods could broaden the characterization of degradation patterns and the corresponding reliability of the selection criteria. This could further refine the practical implications of our findings, for example by identifying validation budgets better suited to the characteristics of some sequences. More broadly, including additional methods would allow future studies to investigate whether particular algorithmic mechanisms tend to produce specific degradation patterns and how these patterns, in turn, relate to the reliability of different selection criteria. Since PPO is the policy-sequence generator considered in the present experiments, characterizing the relationships between algorithmic mechanisms and degradation patterns falls beyond the scope of the present analysis.

\subsection{Improved selection criteria}

The objective of this work was intentionally focused on analyzing commonly used policy selection criteria, rather than proposing new selection mechanisms. The empirical analysis of the \textit{last}, \textit{train}, and \textit{validation} criteria nevertheless highlights limitations that naturally motivate the development of improved selection criteria. In particular, the analysis reveals complementary trade-offs between the reliability of the return estimates used for policy selection and the computational cost required to obtain them.

One direction for improving the \textit{train} criterion is to replace or complement its empirical policy return estimates with more reliable ones based on information collected during training. As discussed in Appendix~\ref{APPENDIX_TrainOverfitting}, degradation can arise when the return estimated during training becomes optimistic relative to the actual performance of the updated policy. Off-Policy Evaluation (OPE)~\citep{thomas_data-efficient_2016} provides a natural direction for addressing this limitation, as it focuses on estimating the performance of a target policy from data collected under previously visited policies, thereby reducing the need for additional environment interaction. OPE has also been used for policy selection in offline RL, where candidate policies must be ranked using previously collected data~\citep{wissmann_point-wise_2026,tang_model_2021,paine_hyperparameter_2020}. However, OPE estimators involve their own trade-offs between estimation bias, variance, and computational cost. For example, importance-sampling-based estimators can exhibit high variance, whereas alternative estimators can reduce variance at the cost of increasing bias~\citep{levine_offline_2020}. These trade-offs would therefore need to be considered when incorporating OPE into the \textit{train} criterion. 

A complementary direction is to reduce the computational cost of the \textit{validation} criterion while preserving its ability to accurately order policies. One possibility is to make validation adaptive both in terms of when policies are evaluated and how much evaluation is performed. Rather than following a fixed validation schedule, event-triggered mechanisms could use information already available during training to identify changes indicative of increased degradation or unreliable training-based return estimates, and trigger additional evaluation only when it is likely to provide useful information~\citep{nguyen_uncertainty-guided_2025}. This would require identifying reliable indicators of degradation that can be monitored from training information without relying on the additional evaluation that they are intended to trigger. The cost of each validation could also be reduced through cheaper or partial performance estimates that allow evaluation to terminate early when sufficient evidence has been obtained to determine that a policy is unlikely to be selected. The checkpointing strategy used in TD7~\citep{fujimoto_for_2023} provides an example of such a mechanism. Its use of the minimum episodic return enables the evaluation of a candidate policy to terminate early when its performance falls below that of the current checkpoint. Such strategies could therefore inspire validation mechanisms that reduce evaluation cost while preserving sufficient information to identify better-performing policies.

\subsection{Selection criteria within the representation of policy sequences}

In practice, sequences of policies are typically represented through learning curves constructed from the return of the policy selected as best throughout the learning process. As observed in our analysis, different selection criteria may identify different policies within the same sequence, meaning that the resulting learning curve can vary depending on the criterion employed. This observation suggests that selection criteria may influence not only the final policy returned by a learning process, but also empirical practices that rely on comparisons between learning curves. Examples include the comparison of RL algorithms~\citep{patterson_empirical_2024,henderson_deep_2018-1} and resource-allocation strategies based on intermediate return estimates~\citep{li_hyperband_2018,bohdal_pasha_2023}.

One illustrative example is the family of early-stopping and adaptive resource-allocation methods, such as successive halving~\citep{jamieson_non-stochastic_2016}, which repeatedly compare the returns of different policy sequences in order to decide how computational resources should be distributed. If selection criteria introduce systematic distortions in the corresponding learning curves, resource-allocation decisions may also be affected. Preliminary experiments exploring this phenomenon are reported in Appendix~\ref{APPENDIX_EarlyStopping}, where initial evidence suggests that default selection criteria implemented in RL libraries may lead to inefficient resource-allocation patterns. A more comprehensive investigation of the influence of selection criteria on fair comparisons between learning curves remains an open research question.

A related topic is the smoothing of learning curves in empirical RL studies, for example through moving averages~\citep{uchendu_jump-start_2023}. Smoothing can facilitate the visualization of overall learning trends and comparisons between algorithms, but it can also obscure the non-monotonic behaviour and performance oscillations of individual policies. This is particularly relevant in the context of policy selection, since smoothing may make a degraded learning process appear more monotonic and can hide performance peaks that determine which policy is actually the best in the sequence. Consequently, a smoothed learning curve may suggest that the \textit{last} policy is the best-performing one, even when individual policies earlier in the sequence achieve higher performance. More generally, as discussed by~\citet{patterson_empirical_2024}, summarizing performance over time into a single representation involves a trade-off between obtaining a more interpretable summary and preserving temporal information. Thus, while smoothing can provide a useful representation of the overall learning trend, it should be interpreted with caution when the objective is to analyze policy-level performance, characterize degradation, or identify the best policy throughout the learning process.

\subsection{Adaptive validation-to-learning cost threshold regulation}

The validation-based criterion highlighted in our analysis stands out for maintaining high accuracy and efficiency across different degradation levels, outperforming the remaining criteria. This criterion relies on a validation-to-learning cost ratio threshold that controls the computational resources invested in policy validation. Our analysis is based on a threshold value of 0.2, for which we observed consistently accurate best-policy tracking across environments and learning regions, while considering an investment of 20\% of the resources devoted to learning in policy validation to be a practical and viable compromise. However, this parameter represents a compromise between tracking accuracy and validation overhead. Larger threshold values allow more policies to be validated, potentially improving tracking accuracy, but at the expense of reducing the computational budget available for learning itself. Conversely, smaller thresholds reduce validation costs but may increase the risk of overlooking high-performing policies. Thus, we consider that studying this trade-off could contribute to the efficient experimental protocols in RL.

In particular, an interesting open question concerns the minimum validation-to-learning cost ratio required to maintain accurate best-policy tracking while reducing validation effort. A related question can be formulated when policy sequences are compared rather than tracked individually. In such settings, the objective is not only to identify the best policy within a sequence, but also to preserve the reliability of comparisons across different learning processes. This introduces a trade-off between validation cost and comparison accuracy, suggesting the possibility of adaptive validation mechanisms that dynamically regulate the validation-to-learning cost threshold while maintaining sufficiently accurate comparisons. Preliminary evidence related to this question is presented in Appendix~\ref{APPENDIX_EarlyStopping}. Moreover, adaptive strategies for controlling such kind of trade-offs have proven useful in other areas of optimization with evolutionary algorithms~\citep{echevarrieta_speeding-up_2024}. Therefore, extending these ideas to interaction-based policy optimization constitutes a promising direction for future research.

\section{Conclusions}\label{SECTION_Conclusions}

In this work, we studied the implications of non-monotonic learning dynamics for policy selection in single-task policy-based RL. Rather than treating the policy returned at the end of training as an intrinsic outcome of the learning process, we explicitly framed it as the result of a selection criterion operating over the sequence of visited policies generated during learning whose true performance may oscillate significantly throughout learning. Within this perspective, we formalized commonly used selection criteria adopted in experimental practice and analyzed their reliability in a representative experimental setting, relying on a well-established on-policy actor-critic method and standard continuous-control benchmark environments, both widely used in the RL literature.

Our results show that, although the magnitude of performance degradation observed during learning is often moderate, it increases with the complexity of the environment and can substantially affect the quality of the selected policy. In low-degradation settings, simple criteria such as returning the last visited policy may already provide accurate and efficient selections. However, in more unstable scenarios, commonly used default criteria based on the last policy or training data can deviate considerably from the best policy visited during learning, leading to distorted representations of learning performance and inefficient use of computational resources. In contrast, we observed that validation-based selection regulated through an explicit computational cost constraint provides a more consistent and reliable approximation of the best visited policy while maintaining high selection efficiency. Furthermore, our analysis shows that these effects emerge even in standard experimental setups based on widely used method, benchmark environments, and default library configurations.

Overall, our findings suggest that policy selection should not be regarded as a secondary implementation detail in RL experimentation. Moreover, the contribution of this work is not to introduce a fundamentally new selection mechanism, but to make explicit that RL experimentation already relies on implicit policy selection procedures whose consequences are rarely formalized or systematically analyzed. We hope this work contributes toward more explicit and standardized validation-based selection protocols in RL evaluation, analogous to the role that validation practices play in supervised learning, and encourages future research on the interaction between instability, performance representation, and computational efficiency in RL.

\subsubsection*{Code Availability}

Alongside the paper, we provide the source code in a GitHub repository: \url{https://github.com/JudithEtxebarrieta/BPT-RL}. The repository provides the code required to reproduce the complete experimental study, including the generation of learning processes, the computation of the proposed diagnostic metrics, the application of the policy selection criteria, and the main analyses and supplementary experiments reported in this work. It also includes a tool for generating learning processes with Stable-Baselines3~\citep{raffin_stable-baselines3_2021,raffin_rl_2020} by selecting the learning algorithm, environment, and policy selection criterion used to track the best policy. Therefore, in addition to supporting the reproduction of the results presented in this work, the repository is intended to facilitate the practical use and further development of selection criteria in non-monotonic learning processes.

\subsubsection*{Acknowledgments}
%

This work was supported by the Basque Government through the Research Groups IT1504-22 2022–2025  program, the BERC 2022–2025 program, and the ELKARTEK program under the projects KK-2026/00055 and KK-2026/00087, and by the Spanish MCIN/AEI/10.13039/501100011033 through projects PID2023-149195NB-I00 and PLEC2024-011247, the latter co-funded by the European Union through the European Regional Development Fund (ERDF).

\bibliography{main}
\bibliographystyle{tmlr}


\appendix

\section{Experimental settings}\label{APPENDIX_SetupConfiguration}

To generate the policy sequences used throughout the analysis, we rely on the default settings provided by the RL libraries, modifying only those aspects that could affect fair comparisons. In this way, our influence is limited solely to the selection of the library and the chosen method--environment pair used to generate the processes, since the remaining parameters determining the process described in Algorithm~\ref{ALGORITHM} are entirely specified by the libraries themselves. This includes the maximum execution time $t_{max}$, the interaction length $n$, and the architecture of the network modelling the policy.

We only consider three modifications, aimed at promoting comparable estimates under uncertainty,  balanced computational resources between learning and validation, and homogeneous experimental conditions across environments. First, unlike the default behaviour of the Stable-Baselines3~\citep{raffin_stable-baselines3_2021} library used to generate the learning processes, the two different subsets of initial states employed for the ground truth and for validation in the \textit{validation} criterion remain fixed for each environment and each generated policy sequence, resulting in paired comparisons for both ground truth and validation~\citep{patterson_empirical_2024}. Second, we perform the validation interaction under the same computational conditions as the learning interaction, using the same level of parallelization by keeping the \texttt{n\_envs} parameter value identical to that determined in RL Baselines3 Zoo~\citep{raffin_rl_2020} when creating both learning and validation environments with \texttt{make\_vec\_env} function in Stable-Baselines3. Third, although state normalization is enabled by default for some environments and disabled for others in RL Baselines3 Zoo, we disable normalization in all environments to maintain homogeneous experimental conditions.

\section{Estimation of policy return}\label{APPENDIX_EmpiricalEstimates}

In this appendix, we discuss and compare in detail the different empirical estimators of the policy return considered throughout this work.

\subsection{Ground truth}\label{APPENDIX_EmpiricalEstimates_GroundTruth}

Each policy $\pi$ in the sequence of policies $\Pi_t$ visited during a learning process has associated a random variable that captures the episodic returns of all episodes that can be generated by the policy starting from all possible initial states, i.e., $X=\big\{r(\tau)\colon \tau\sim (s_0,\pi,p),~s_0\in S_I\big\}$. The real objective value of a policy depends on this variable (see Equation~\ref{EQUATION_ExpectedDiscountedCumulativeReward}). However, in practice only a finite sample is observable~\citep{jordan_position_2024}. Therefore, in this work, for each environment considered, we fix a large subset of initial states $S_I^{truth}\subset S_I$ and represent the policy ground truth objective as the empirical estimate (see Equation~\ref{EQUATION_EmpiricalEstimate}) computed from the episodes generated by the policy starting from those initial states, i.e., $D=\big\{\tau\colon \tau\sim (s_0,\pi,p),~s_0\in S_I^{truth}\big\}$. Specifically, we consider a subset of 500 initial states, a size for which we observe stable estimates.

\begin{figure}[H]
\centering
\includegraphics[width=\textwidth]{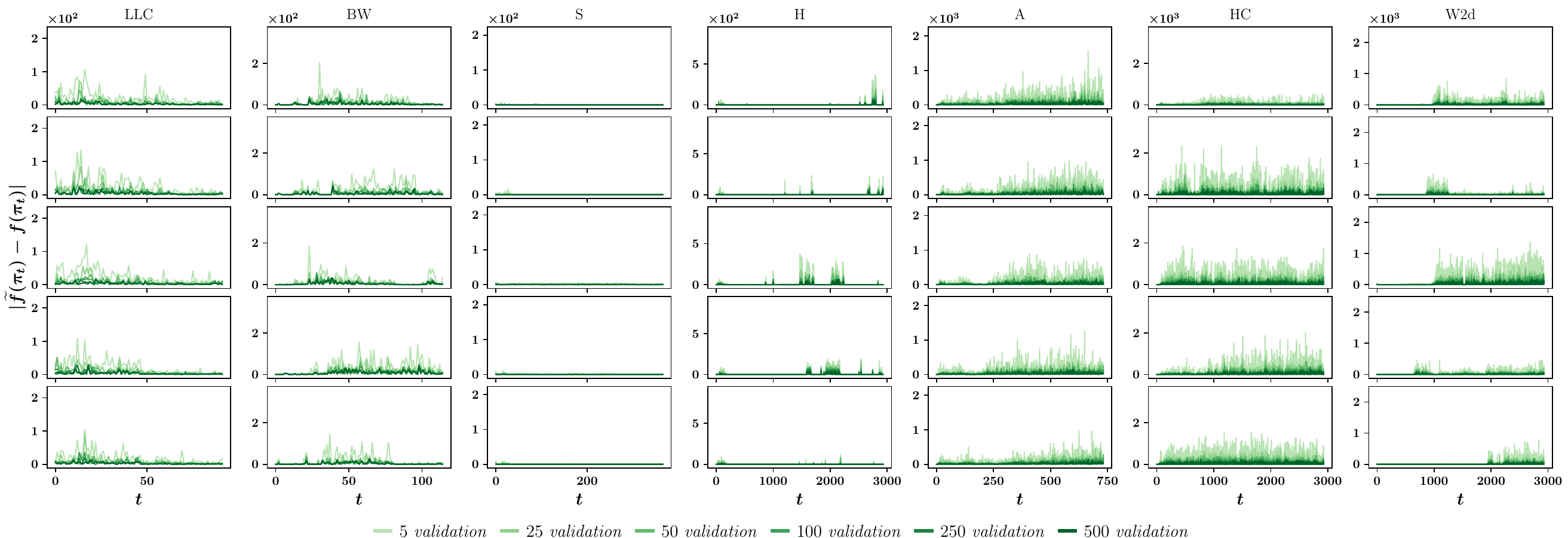}
\caption{Evolution of the absolute differences between the ground truth and the empirical estimators of \textit{validation} criteria across the policies in the generated sequences for each environment (by column) and for 5 different seeds (by row). Six curves are shown, corresponding to different numbers of episodes $\kappa$ used for the estimation, where darker curves indicate a larger number of episodes considered.}\label{FIGURE_GroundTruth}
\end{figure}

In Figure~\ref{FIGURE_GroundTruth}, different darknesses of green show the evolution of the difference between the ground truth empirical estimates obtained from 500 initial states and those obtained from another independent subset of initial states of varying size. We observe that, as the size of the second subset approaches 500, the corresponding curve becomes closer to the horizontal line at 0, and this behavior already appears for substantially smaller subset sizes. Notice that only the lightest green curves visibly deviate from this horizontal line. This suggests that even using a different and smaller sample of initial states produces very similar estimates, supporting that the 500 episodes considered for the ground truth are sufficient.

\begin{table}[H]
\centering
\caption{Spearman rank correlation between policy rankings obtained using 500 initial states and those obtained using smaller independent subsets of initial states, averaged across the 30 policy sequences for each environment. Values are reported as mean $\pm$ standard deviation. Gray cells indicate the smallest subset size for which the mean correlation minus one standard deviation exceeds 0.95.}\label{TABLE_EstimatesOrdering}
{\footnotesize
\begin{tabular}{|r|c|c|c|c|c|c|c|}
\hline
$|D|$ & LLC & BW & S & H & A & HC & W2d \\
\hline
5 & 0.835 ± 0.063 & 0.892 ± 0.063 & 0.974 ± 0.030 & \cellcolor{gray!30}0.978 ± 0.021 & 0.922 ± 0.037 & 0.898 ± 0.068 & 0.952 ± 0.034 \\
25 & 0.927 ± 0.039 & 0.944 ± 0.033 & \cellcolor{gray!30}0.992 ± 0.011 & 0.992 ± 0.007 & \cellcolor{gray!30}0.978 ± 0.011 & 0.969 ± 0.028 & \cellcolor{gray!30}0.986 ± 0.014 \\
50 & 0.950 ± 0.027 & \cellcolor{gray!30}0.961 ± 0.025 & 0.997 ± 0.004 & 0.996 ± 0.004 & 0.988 ± 0.007 & \cellcolor{gray!30}0.983 ± 0.016 & 0.992 ± 0.008 \\
100 & 0.966 ± 0.023 & 0.975 ± 0.018 & 0.999 ± 0.002 & 0.997 ± 0.003 & 0.993 ± 0.004 & 0.990 ± 0.010 & 0.995 ± 0.005 \\
250 & \cellcolor{gray!30}0.980 ± 0.014 & 0.984 ± 0.011 & 0.998 ± 0.003 & 0.999 ± 0.002 & 0.996 ± 0.002 & 0.995 ± 0.005 & 0.998 ± 0.003 \\
500 & 0.986 ± 0.010 & 0.987 ± 0.009 & 0.999 ± 0.002 & 0.999 ± 0.001 & 0.997 ± 0.001 & 0.996 ± 0.004 & 0.998 ± 0.002 \\
\hline
\end{tabular}
}
\end{table}

While the previous analysis evaluates the stability of the estimated performance of individual policies, the reliability of the estimator is also relevant for preserving the relative ordering of the policies in the sequence. To assess this aspect, Table~\ref{TABLE_EstimatesOrdering} reports the Spearman rank correlation between the policy ranking obtained using the 500 initial states considered for the ground truth and the ranking obtained using an independent subset of initial states of varying size. A higher correlation indicates that the ordering induced by the smaller subset is closer to the ordering obtained with the 500 initial state reference estimate. The results show that the rankings are already highly correlated for sample sizes substantially smaller than 500. In particular, in all considered environments, the mean correlation minus one standard deviation exceeds 0.95 for a subset size smaller than 500, with the corresponding minimum sample sizes highlighted in gray. This provides additional empirical evidence that the sample size used for the ground truth yields stable policy orderings and that the relative ordering of the policies is not substantially affected by the finite-sample variability of the performance estimates in the considered experimental setting.

\subsection{Estimates by \textit{train} and \textit{validation} criteria}\label{APPENDIX_TrainTestEstimates}
The \textit{train} and \textit{validation} criteria defined in Section~\ref{SECTION_SelectionCriteria} also rely on empirical estimates of the policy return. However, the set $D$ used in these criteria differs from the one considered for the ground truth in several crucial aspects, which makes their estimates inherently less reliable.

\begin{figure}[H]
\centering
\includegraphics[width=\textwidth]{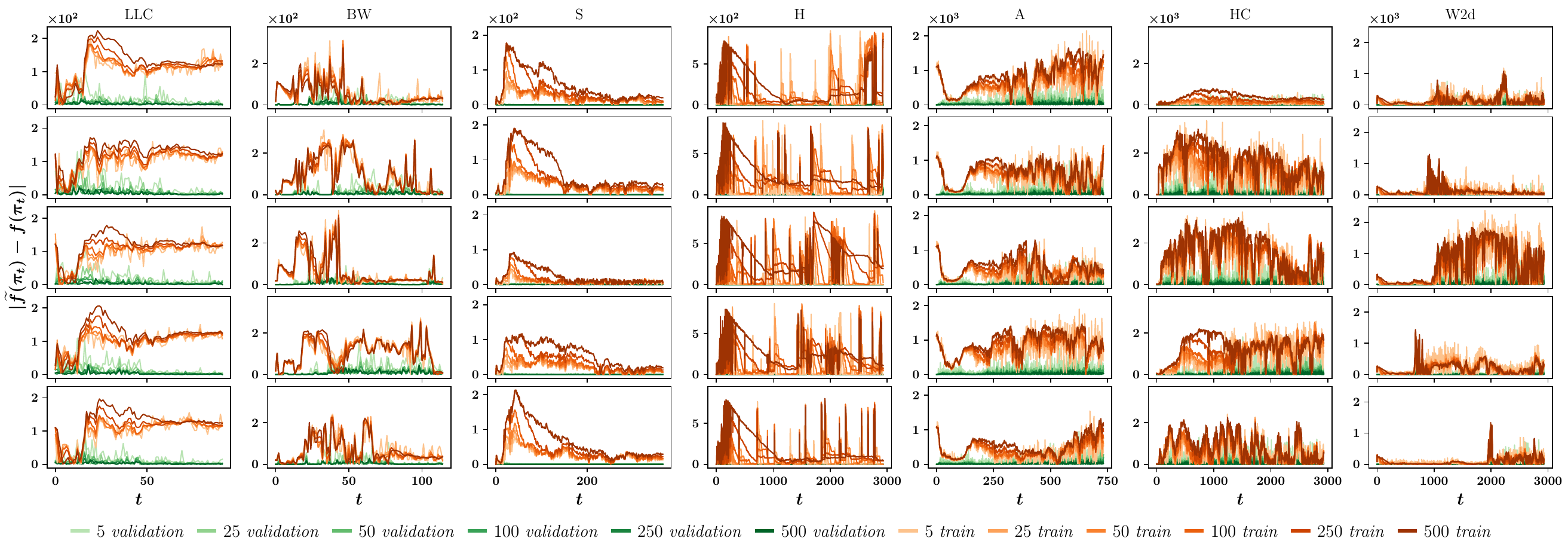}
\caption{Evolution of the absolute differences between the ground truth and the empirical estimators of \textit{train} (orange) and \textit{validation} (green) criteria across the policies in the generated sequences for each environment (by column) and for 5 different seeds (by row). Six curves are shown per criterion, corresponding to different numbers of episodes $\kappa$ used for the estimation, where darker curves indicate a larger number of episodes considered. The \textit{validation} curves coincide with those shown in Figure~\ref{FIGURE_GroundTruth}.}\label{FIGURE_EmpiricalEstimates}
\end{figure}

The \textit{train} criterion reuses episodes generated during learning, resulting from the concatenation of trajectories generated by the policies in the sequence. Consequently, the estimate associated with each policy may include episodes generated by different policies. Moreover, estimates across policies are not paired, since they rely on episodes initialized from different initial states. In addition, the episodes are generated under the stochasticity injected during training to encourage exploration, which is absent during validation. All these factors contribute to the \textit{train} estimates being substantially worse than those of \textit{validation}, as illustrated in Figure~\ref{FIGURE_EmpiricalEstimates}. The orange curves lie farther from the horizontal line at 0 than the green curves. In some cases, such as LLC and S, considering more episodes even increases the gap, since the darker orange curves deviate more strongly. This is consistent with the effect of incorporating too many episodes generated by policies that are increasingly different from the current one.

For the \textit{validation} criterion, the situation improves considerably. The estimates become unbiased again because the validation episodes are generated by the evaluated policies themselves. Furthermore, comparisons are paired, since the same subset of initial states is considered when validating different policies. As a result, even the curves associated with a very small number of validation episodes remain much closer to the horizontal line at 0 than any of the \textit{train} curves. It is worth noting that the last desirable property is not satisfied in some libraries such as Stable-Baselines3~\citep{raffin_stable-baselines3_2021}, where, by default, validation episodes are generated from newly sampled initial states each time a policy is evaluated, potentially producing different initial state samples across validations.

\subsection{Overfitting of \textit{train} estimates}\label{APPENDIX_TrainOverfitting}

It is observed in the RL literature that return estimates based on training data may fail to generalize to the true underlying policy return, a phenomenon commonly referred to as overfitting~\citep{zhang_study_2018}. This issue is particularly relevant for the \textit{train} criterion, where the same data used to update the policy is also used to construct empirical return estimates, potentially introducing a shared bias. 

\begin{figure}[H]
\centering
\includegraphics[width=\textwidth]{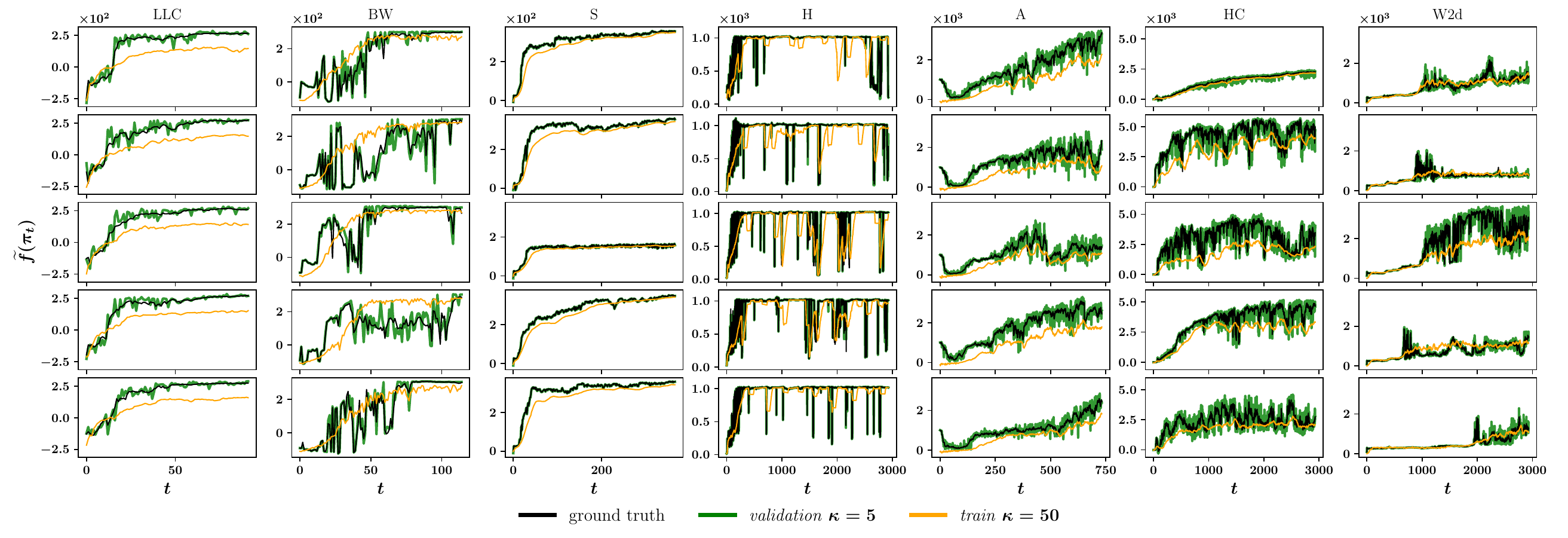}
\caption{Evolution of the ground-truth episodic return (black curve) and empirical return estimates from training trajectories (orange curve for the \textit{train} criterion) and validation episodes (green curve for the \textit{validation} criterion) across five different policy sequences for each environment. }\label{FIGURE_TrainOverfitting}
\end{figure}

In this appendix, we further illustrate this effect by analyzing the evolution of empirical return estimates along five policy sequences for each environment. Figure~\ref{FIGURE_TrainOverfitting} compares, for each sequence, three curves: the ground-truth return, the estimate computed from training episodes using the selected \textit{train} configuration, and the estimate computed from validation episodes using the selected \textit{validation} configuration. These configurations correspond to those used in the main experimental analysis in Section~\ref{SECTION_ExperimentalAnalysis}, which were selected due to their overall high accuracy across environments and learning regions (see Appendix~\ref{APPENDIX_Configurations_Experiments} for further details).

The results highlight clear differences between the two estimators. The \textit{validation} estimate closely follows the ground-truth curve, accurately capturing non-monotonic behaviour and return drops throughout training. In contrast, the \textit{train} estimate is generally smoother and often exhibits an overly monotonic increasing trend. In some cases, it partially captures fluctuations in return, but these are attenuated or detected with a noticeable delay. This behaviour is consistent with the use of shared training data for both policy updates and return estimation, which can propagate bias from the learning dynamics into the estimator itself. Overall, these results provide further evidence that validation-based estimates are more reliable for tracking true policy return during learning.

\section{Selection criteria within on-policy learning processes}\label{APPENDIX_SelectionCrietriaInPractice}
In Section~\ref{SECTION_TheoreticalFramework}, the interaction-based STPO learning process and the selection criteria were defined to formalize the conditions for policy selection and to specify the policy ultimately returned. These definitions make explicit the policy that is returned in practice from the sequence of visited policies. However, in practical implementations, learning processes do not store the entire sequence of visited policies, as represented in line~9 of Algorithm~\ref{ALGORITHM}. Instead, they only track the best policy observed so far, which becomes the only policy that is effectively stored during learning.

In this appendix, we explicitly illustrate how selection criteria operate in practice during learning processes. We present one algorithm for each criterion, all of them based on the same interaction-based learning framework described in Algorithm~\ref{ALGORITHM}, but differing in how the best solution is tracked throughout learning. In particular, we make explicit the operations hidden behind line~6 of Algorithm~~\ref{ALGORITHM}, which are now highlighted in different colors, while also showing the actual data storage performed during learning.

It is important to note that the policies returned by the practical Algorithms~\ref{ALGORITHM_last}, \ref{ALGORITHM_train}, and \ref{ALGORITHM_test} coincide with the policies defined by the selection criteria in Equations~\ref{EQUATION_last}, \ref{EQUATION_train}, and \ref{EQUATION_test}, respectively. The difference lies in how they are represented: in the theoretical framework, the criteria are defined explicitly over the complete sequence of visited policies, whereas practical implementations only keep track of the best policy observed so far during learning.

\begin{minipage}[b]{0.48\textwidth}

\begin{algorithm}[H]
\caption{On-policy STPO with {\color{blue}\textit{last}}}\label{ALGORITHM_last}
{\small
\begin{algorithmic}[1]

\Statex \hspace*{-1.5em} \parbox[t]{1\linewidth}{  \textbf{Input:}\\ 
$\mathcal{E}$: Environment;\\
 $\rho$: Random seed; \\
$t_{max}$: Maximum learning interaction steps;
}
\vspace{2mm}
\Statex \hspace*{-1.5em}  \textbf{Output:} Learned policy
\vspace{2mm}

\State $t\gets 0$
\State $\pi_0\gets$ Initialize policy with $\rho$

\While{$t < t_{max}$}
    \State  $\tau_t\gets$ Interaction of $\pi_t$ with $\mathcal{E}$ during $n$ steps\\[-4mm]
    \tikzmk{A}\State$\widetilde{\pi}_t\gets \pi_t$\tikzmk{B}\boxitStateIn{myblue}
\State $\pi_t\gets$ Update $\pi_t$ using $\tau_t$
\State $t\gets t+ n$

\EndWhile

\State \Return $\widetilde{\pi}_t$

\end{algorithmic}
}
\end{algorithm}

\end{minipage}
\hfill
\begin{minipage}[b]{0.48\textwidth}
Our experimental analysis is based on Proximal Policy Optimization~\citep{schulman_proximal_2017-1}, which is an interaction-based and, more specifically, an on-policy actor-critic method. On-policy methods update the policy exclusively using trajectories generated by the current policy itself (lines~6, 11, 12, and 17 of Algorithms~\ref{ALGORITHM_last}, \ref{ALGORITHM_train}, \ref{ALGORITHM_test} and \ref{ALGORITHM_CostDrivenTest}, respectively). Consequently, in practice, these methods do not need to store all trajectories generated during learning either, meaning that line~5 of Algorithm~\ref{ALGORITHM} is not required for the policy update itself.

Algorithm~\ref{ALGORITHM_last}, corresponding to the \textit{last} criterion, simply keeps the current policy at each iteration, without storing additional policies or learning trajectories. 

\end{minipage}

Algorithm~\ref{ALGORITHM_train}, associated with the \textit{train} criterion, instead depends on an additional input parameter $\kappa$ and requires storing learning trajectories, since they are used to compute the empirical estimate of the visited policies (lines~6--7). In this case, the stored best policy is updated whenever the current policy achieves a higher empirical estimate than the best one observed so far (lines~8--10).

\begin{minipage}[b]{0.48\textwidth}

\begin{algorithm}[H]
{\small
\caption{On-policy STPO with {\color{orange}\textit{train}}}\label{ALGORITHM_train}
\begin{algorithmic}[1]

\Statex \hspace*{-1.5em} \parbox[t]{1\linewidth}{  \textbf{Input:}\\ 
$\mathcal{E}$: Environment;\\
 $\rho$: Random seed; \\
$t_{max}$: Maximum learning interaction steps;\\
\tikzmk{A}
$\kappa$: Number of validation episodes;\tikzmk{B}\boxitInput{myorange}

}

\Statex
\Statex \hspace*{-1.5em}  \textbf{Output:} Learned policy
\vspace*{1mm}

\Statex 
\State $t\gets 0$
\State $\pi_0\gets$ Initialize policy with $\rho$
\tikzmk{A}
\State $D_t\gets \emptyset$, $\widetilde{f}_{max}\gets -\infty$
\tikzmk{B}\boxitState{myorange}

\While{$t < t_{max}$}

    \State  $\tau_t\gets$ Interaction of $\pi_t$ with $\mathcal{E}$ during $n$ steps\\[-2mm]
    \tikzmk{A}\State $D_t\gets D_t\cup \{\tau_t\}$
    \State$D \gets$ Last $\kappa$ episodes in $D_t$\\[-2mm]
    
    \If{$\widetilde{f}(\pi_t;D)>\widetilde{f}_{max}$}
        \State$\widetilde{f}_{max}\gets \widetilde{f}(\pi_t;D)$
        \State$\widetilde{\pi}_t\gets \pi_t$

\EndIf \tikzmk{B}\boxit{myorange}

\State $\pi_t\gets$ Update $\pi_t$ using $\tau_t$
\State $t\gets t+ n$

\EndWhile

\State \Return $\widetilde{\pi}_t$

\end{algorithmic}
}
\end{algorithm}

\end{minipage}
\hfill
\begin{minipage}[b]{0.48\textwidth}

\begin{algorithm}[H]
\caption{On-policy STPO with {\color{purple}\textit{validation}}}\label{ALGORITHM_test}
{\small
\begin{algorithmic}[1]

\Statex \hspace*{-1.5em} \parbox[t]{1\linewidth}{  \textbf{Input:}\\ 
$\mathcal{E}$: Environment;\\
 $\rho$: Random seed; \\
$t_{max}$: Maximum learning interaction steps;\\
\tikzmk{A}
$\varphi$: Validation period;\\
$\kappa$: Number of validation episodes;
\tikzmk{B}\boxitInput{mypurple}
}

\Statex \hspace*{-1.5em}  \textbf{Output:} Learned policy

\State $t\gets 0$
\State $\pi_0\gets$ Initialize policy with $\rho$
\tikzmk{A}
\State $\widetilde{f}_{max}\gets -\infty$
\State $S_I^{test}\gets$ Drawn $\kappa$ initial states from $p_I $
\tikzmk{B}\boxitState{mypurple}

\While{$t < t_{max}$}

    \State  $\tau_t\gets$ Interaction of $\pi_t$ with $\mathcal{E}$ during $n$ steps\\[-2mm]
  \tikzmk{A}\If{$\varphi \mid t$}

    \State$D \gets \big\{ \tau\colon \tau\sim (s_0,\pi_t,p),~ s_0\in S_I^{test}\big\}$
    
    \If{$\widetilde{f}(\pi_t;D)>\widetilde{f}_{max}$}
        \State$\widetilde{f}_{max}\gets \widetilde{f}(\pi_t;D)$
        \State$\widetilde{\pi}_t\gets \pi_t$
    \EndIf
\EndIf \tikzmk{B}\boxit{mypurple}

\State $\pi_t\gets$ Update $\pi_t$ using $\tau_t$
\State $t\gets t+ n$

\EndWhile

\State \Return $\widetilde{\pi}_t$

\end{algorithmic}
}
\end{algorithm}
\end{minipage}

Algorithm~\ref{ALGORITHM_test} represents the \textit{validation} criterion. This criterion introduces an additional input parameter, the validation period $\varphi$, which determines how often validation is performed. Unlike \textit{train}, this criterion does not store learning trajectories. Instead, whenever the number of interaction steps elapsed becomes a multiple of the validation period $\varphi$ (line~7), validation episodes generated by the current policy are collected (line~8) starting from the $\kappa$ validation initial states fixed at the beginning of learning (line~4). Afterwards, the same update operation used in \textit{train} is performed to replace the currently stored best policy whenever a higher empirical estimate is obtained, now using validation episodes to compute the estimate.

\begin{minipage}[b]{0.48\textwidth}

\begin{algorithm}[H]
\caption{\\On-policy STPO with {\color{green!50!black}cost-drive \textit{validation}}}\label{ALGORITHM_CostDrivenTest}
{\small
\begin{algorithmic}[1]

\Statex \hspace*{-1.5em} \parbox[t]{1\linewidth}{  \textbf{Input:}\\ 
$\mathcal{E}$: Environment;\\
 $\rho$: Random seed; \\
$t_{max}$: Maximum learning interaction steps;\\
\tikzmk{A}
$\kappa$: Number of validation episodes;\\
$\varphi_c$: Validation-tolearning cost ratio threshold;\tikzmk{B}\boxitInput{mygreen}
}

\Statex \hspace*{-1.5em}  \textbf{Output:} Learned policy

\State $t\gets 0$
\State $\pi_0\gets$ Initialize policy with $\rho$
\tikzmk{A}
\State $\widetilde{f}_{max}\gets -\infty$
\State  $c_{learn}\gets 0$, $c_{val}\gets 0$, $\widetilde{c}\gets 0$, $c_\tau\gets -\infty$
\State $S_I^{test}\gets$ Drawn $\kappa$ initial states from $p_I $
\tikzmk{B}\boxitState{mygreen}

\While{$t < t_{max}$}

    \State  $\tau_t\gets$ Interaction of $\pi_t$ with $\mathcal{E}$ during $n$ steps\\[-2mm]
    \tikzmk{A}\State $c_{learn}\gets c_{learn} + c(\tau_t)+c(\pi_t)$
    
    \If{$\dfrac{c_{val}+\widetilde{c}}{c_{learn}}\leq \varphi_c$}
    \State$D \gets \{\tau\colon \tau\sim (s_0,\pi_t,p),s_0\in S_I^{test}\}$ 
    
    \If{$\widetilde{f}(\pi_t;D)>\widetilde{f}_{max}$}
        \State$\widetilde{f}_{max}\gets \widetilde{f}(\pi_t;D)$
        \State$\widetilde{\pi}_t\gets \pi_t$
      \EndIf
      
     \State $c_\tau\gets \max \Big\{ c_\tau, \underset{\tau\in D}{\max}~c(\tau)\Big\}$
     \State $\widetilde{c}\gets  c_\tau\cdot \kappa$
     \State $c_{val}\gets c_{val}+ \underset{\tau\in D}{\sum}~c(\tau)$

\EndIf \tikzmk{B}\boxit{mygreen}

\State $\pi_t\gets$ Update $\pi_t$ using $\tau_t$
\State $t\gets t+ n$

\EndWhile

\State \Return $\widetilde{\pi}_t$

\end{algorithmic}
}
\end{algorithm}

\end{minipage}
\hfill
\begin{minipage}[b]{0.48\textwidth}

Finally, Algorithm~\ref{ALGORITHM_CostDrivenTest} defines the proposed slightly modified version of the original \textit{validation} criterion. This criterion is guided by computational cost, rather than by the number of interaction steps $t$ that limits the learning process. During learning, three variables associated with computational cost are updated: the estimated validation cost of $\kappa$ episodes $\widetilde{c}$, the total cost invested in validation $c_{val}$, and the total cost invested in learning $c_{learn}$ (initialized in line~4). At each iteration, the learning cost is increased by the cost of the interaction and the policy update (line~8), while the validation cost is increased whenever $\kappa$ validation episodes are generated (line~16). Consequently, at each iteration, a check is performed to determine whether validating the current policy is expected to incur a computational cost lower than or equal to the predefined threshold (line~9). If this condition holds, the policy is validated (lines~10--13), and the validation cost estimate is updated using a pessimistic estimate based on the maximum validation episode cost observed so far (lines~14--15).

The notation $c(\cdot)$ used in lines~8, 14, and 16 denotes the computational cost required to obtain the corresponding element. For example, $c(\tau_t)$ and $c(\pi_t)$ represent the cost of performing the last interaction of $n$ steps and the last policy update in lines~7 and~17, respectively.

\end{minipage}

\section{Configuration of \textit{train} and \textit{validation} criteria}\label{APPENDIX_CriteriaConfiguration}

In this appendix, we explicitly describe the selection of the \textit{train} and \textit{validation} criteria configurations considered in the experiments of Section~\ref{SECTION_ExperimentalAnalysis}. In addition, we include an analysis of the validation cost associated with the different policies in the sequences of policies generated during a learning process, and how this influences the validation-to-learning cost ratio for both the default \textit{validation} criterion and the cost-driven \textit{validation} criterion proposed in this work.

\subsection{Criteria configuration for experimentation}\label{APPENDIX_Configurations_Experiments}
In Stable-Baselines3 library~\citep{raffin_stable-baselines3_2021}, several default values are defined that determine the configuration of the \textit{train} and \textit{validation} criteria. On the one hand, the parameter \texttt{stats\_window\_size}, by default set to 100 for any environment within the \texttt{PPO} class associated with the on-policy Proximal Policy Optimization method~\citep{schulman_proximal_2017-1}, plays the role of the parameter $\kappa$ defined in this paper for the \textit{train} criterion, as the number of training episodes considered for the empirical estimation. On the other hand, in the \texttt{EvalCallback} class, which allows periodic validation of the visited policies, the parameters \texttt{n\_eval\_episodes} and \texttt{eval\_freq}, with default values of 5 and 10000 respectively, play the role of $\kappa$ and $\varphi$ defined in this paper for the \textit{validation} criterion. In this latter case, the evaluation period \texttt{eval\_freq} is measured in interaction steps, which corresponds to different numbers of iterations (see column 4 of Table~\ref{TABLE_configurations}), since for each environment RL Baselines3 Zoo library~\citep{raffin_rl_2020} uses different learning interaction lengths (represented by $n$ in Algorithm~\ref{ALGORITHM}). These default configurations are used for the \textit{train} and \textit{validation} criteria in the Figures~\ref{FIGURE_SH_LearningCurves} and \ref{FIGURE_ResourceAllocation} of Appendix~\ref{APPENDIX_EarlyStopping}.

\begin{table}[H]
\centering
\caption{Default and selected configurations for the \textit{train} and \textit{validation} criteria across environments.}\label{TABLE_configurations}
{\scriptsize
\begin{tabular}{|c|c|c|c|c|c|}
\hline 
$\mathcal{E}$ & \thead{\textit{Train} default\\$\kappa$} & \thead{\textit{Validation} default\\$\kappa$} & \thead{\textit{Validation} default $\varphi$\\10000 steps $\approx$ iter.} & \thead{\textit{Train} selected\\$\kappa$} & \thead{Cost-driven \textit{validation} selected\\$(\kappa,\varphi_c)$} \\ 
\hline 
LLC  & \multirow{7}{*}{100} & \multirow{7}{*}{5} & 10 & \multirow{7}{*}{50} & \multirow{7}{*}{(5,0.2)} \\ 
\cline{1-1}\cline{4-4}
BW   &  &  &  5 &  &  \\ 
\cline{1-1}\cline{4-4}
S    &  &  & 10 &  &  \\ 
\cline{1-1}\cline{4-4}
H    &  &  & 20 &  &  \\ 
\cline{1-1}\cline{4-4}
A    &  &  & 5  &  &  \\ 
\cline{1-1}\cline{4-4}
HC   &  &  & 20 &  &  \\ 
\cline{1-1}\cline{4-4}
W2d  &  &  & 20 &  &  \\

\hline 
\end{tabular} 
}
\end{table}
\vspace{-6mm}
\begin{figure}[H]
\centering
\includegraphics[width=\textwidth]{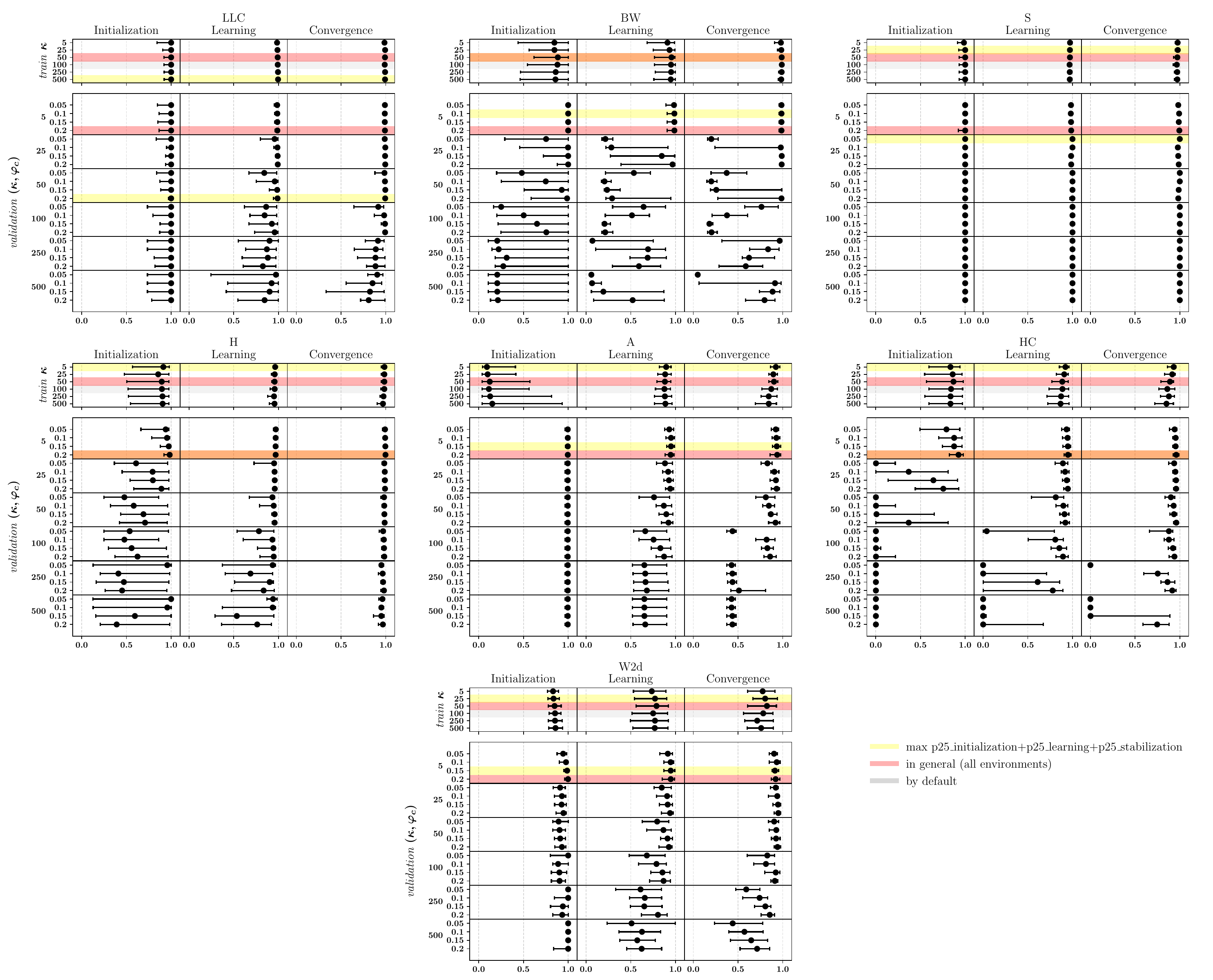}
\vspace{-6mm}
\caption{Distribution of selection accuracy for different configurations of the \textit{train} and cost-driven \textit{validation} criteria across environments and regions. Each distribution is represented by its interquartile range, with the median marked within it. The most accurate configurations overall across all regions and environments are highlighted in red, those that are most accurate for each individual environment are highlighted in yellow, and the default configurations are highlighted in grey.}\label{FIGURE_Configurations}

\end{figure}

For the main analysis of the criteria in Section~\ref{SECTION_ExperimentalAnalysis} instead, we fix configurations for the \textit{train} and cost-driven \textit{validation} criteria that perform well across all environments. To this end, we define a grid of possible parameter values and analyze the selection accuracies (see Equation~\ref{EQUATION_accuracy}) of each configuration across the three regions. Specifically, we consider $\kappa \in \{5,25,50,100,250,500\}$ and $\varphi_c \in \{0.05,0.1,0.15,0.2\}$, and compare the distribution of accuracy associated with each configuration across regions and environments, as shown in Figure~\ref{FIGURE_Configurations}. Finally, we select the configuration with the highest sum of 25th percentiles across all regions and environments, which we highlight in red and specify in the last columns of Table~\ref{TABLE_configurations}. In Figure~\ref{FIGURE_Configurations}, we also highlight in grey the default \textit{train} configuration, and in yellow the best configurations when considering the three regions of each environment individually. Overall, there are no significant differences between the segments of the best configurations per environment and the globally selected one, which indicates consistency and low sensitivity of selection accuracy across environments for the selected configurations.

Figure~\ref{FIGURE_Configurations} also reveals several interesting patterns. First, the \textit{train} criterion is less sensitive to parameter changes than the cost-driven \textit{validation} criterion, with segments across the three regions varying very little for different values of $\kappa$. However, some deviation is observed in the learning and convergence regions as $\kappa$ increases, as expected due to the inclusion of more episodes generated by policies further away from the current one in the empirical estimation (see Equation~\ref{EQUATION_EmpiricalEstimate}). This deviation is more pronounced for the cost-driven \textit{validation} criterion. In this case instead, the reason is not that more validation episodes lead to higher inaccuracies, but rather that a larger number of episodes increases the validation cost, so that for a fixed validation-to-learning cost threshold fewer policies are evaluated, with longer periods between validations that more easily overlook good policies.

\vspace{-2mm}
\begin{minipage}[b]{0.48\textwidth}

\subsection{Validation cost within \textit{validation} criteria}\label{APPENDIX_Validation_Cost}
\vspace{2mm}
Given a sequence of policies visited during a learning process, validation with the same number of episodes for different policies does not necessarily imply the same cost. This makes it difficult to define a common \textit{validation} configuration that ensures a similar cost across all cases.

\subsubsection{Irregular validation cost within sequences}
\vspace{2mm}

Figure~\ref{FIGURE_val_cost_seq} shows the validation-to-learning cost ratio associated with each individual policy in a sequence for each environment. The last six rows of each plot illustrate these cost ratios for different numbers of episodes, with a color scale ranging from dark green for ratios close to 0, to lighter green and white as the ratio approaches 0.3, and from white to red as it reaches 1, with ratios clipped at 1. Thus, when the validation cost of a policy is less than 30\% of the time spent on its learning interaction and update, it is shown in green (darker green meaning lower cost), whereas higher costs are shown in red (darker red meaning higher cost). In addition, the first two rows represent, in grayscale, the normalized ground truth and normalized average episode length of each policy throughout the sequence. Darker values indicate higher ground truth and longer episodes.

\vspace{2.2cm}

\end{minipage}
\hfill
\begin{minipage}[b]{0.48\textwidth}
\begin{figure}[H]
\centering
\includegraphics[width=\textwidth]{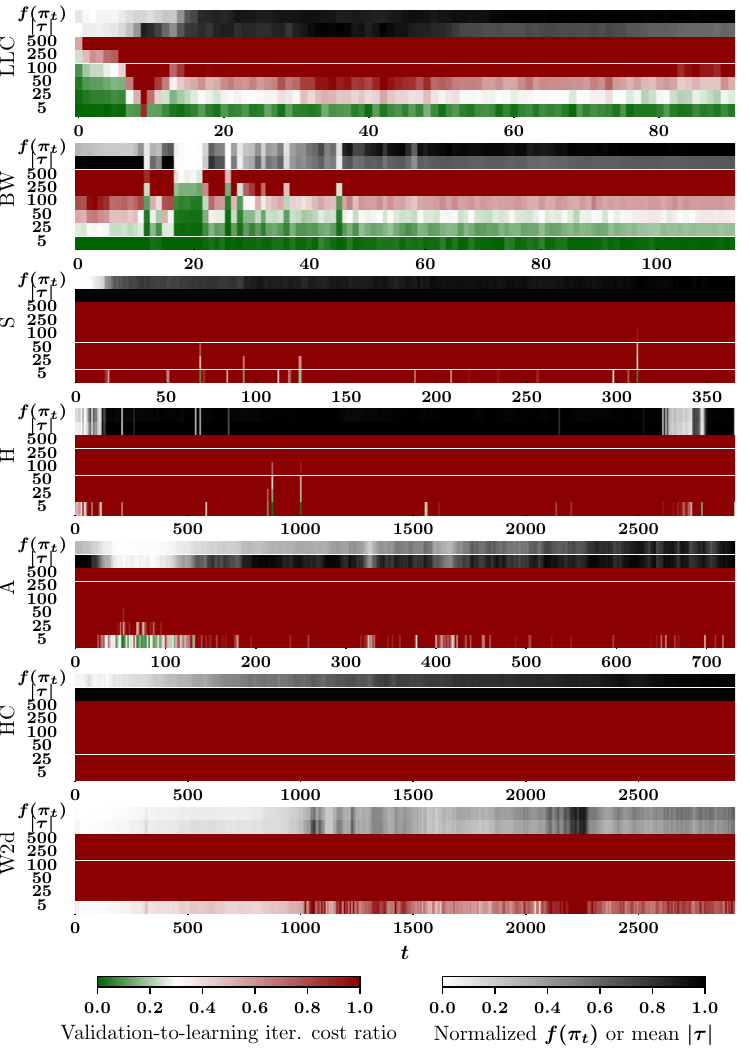}
\caption{In each plot, the ground truth (first row), the average length of generated episodes (second row), and the validation-to-learning cost ratio per iteration (last six rows) for the policies in a sequence generated under each environment.}\label{FIGURE_val_cost_seq}
\end{figure}
\end{minipage}

It can be observed that the rows are not homogeneous in color, meaning that validating different policies leads to different costs. Moreover, this pattern is not consistent across environments, suggesting that validation cost also varies across environments. Furthermore, the smaller the number of episodes used for validation, the lower the validation-to-learning cost ratio. Notice that the lowest rows show clearer and greener patterns.

It is worth noting that, across each sequence, for all environments except S and HC, the policies with lower cost ratios coincide with those with lower ground truth, which in turn correspond to those generating shorter episodes. These regions are typically located at the beginning of the sequence, indicating that worse policies are cheaper to validate because they generate shorter episodes. In contrast, for S and HC, all policies generate episodes of the same length, which makes validation cost more homogeneous. This is because in S and HC episode duration is fixed to 1000 interaction steps, whereas in the other environments episodes may terminate earlier upon reaching an unhealthy state, which is a terminal state from which no further interaction is possible, e.g., in A or W2d when the torso coordinates are very close to the ground, indicating that either the ant is upside down or the biped has fallen over, respectively.

\subsubsection{Validation cost across different criteria configurations}
\begin{figure}[H]
\centering
\includegraphics[width=\textwidth]{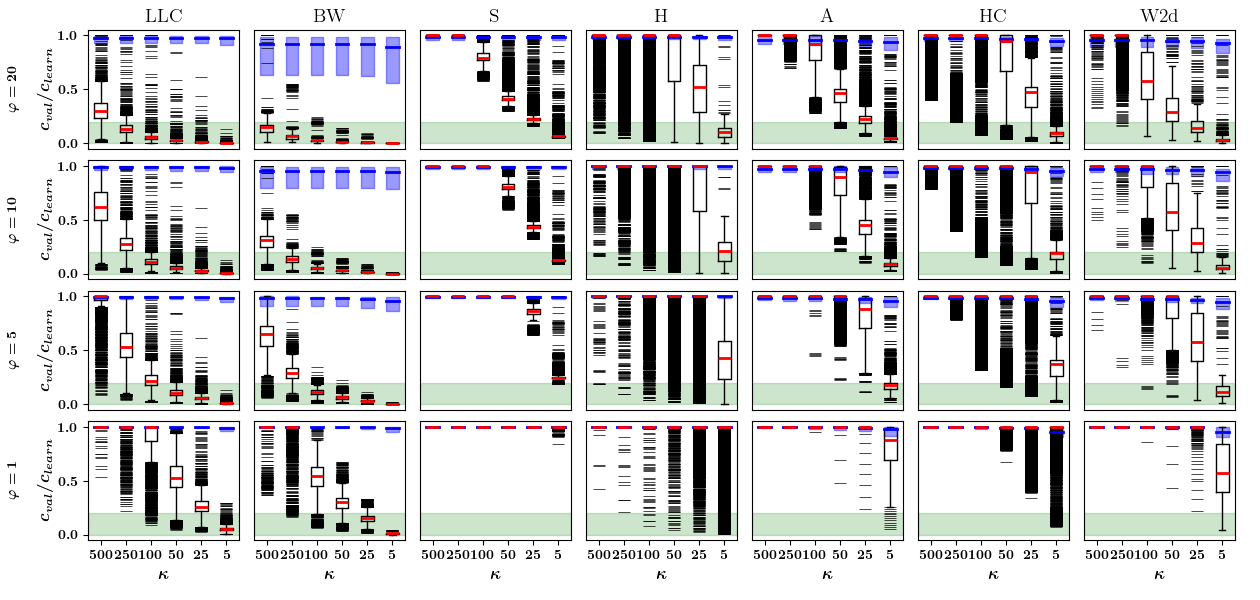}
\vspace{-7mm}
\caption{Boxplots of the validation-to-learning cost ratio for the default \textit{validation} criterion across the 30 generated learning processes, considering different validation periods (rows) and numbers of validation episodes (columns within each plot), for each environment (columns). The green band highlights the admissible validation-to-learning cost ratio range between 0 and 0.2, while the blue boxes in the corresponding columns show the interquartile range and median of the selection accuracy.}\label{FIGURE_cost_test_default_boxplots}

\end{figure}

This irregularity in validation cost across and within policy sequences is reflected in the distribution of validation costs for different configurations of the \textit{validation} criterion. Figure~\ref{FIGURE_cost_test_default_boxplots} shows a boxplot of the validation-to-learning cost ratio across learning processes defined by each different configuration. It can be observed that, except for the simpler environments LLC and BW, most boxes lie above the cost ratio threshold of 0.2 illustrated by the green band. This highlights the difficulty of selecting a configuration with controlled validation cost. From the entire grid analysis of different possible configurations, it can be concluded that for a larger validation period $\varphi=20$, the validation cost becomes more manageable, especially when combined with a small number of validation episodes such as $\kappa=5$. However, increasing the validation period affects selection accuracy, which is illustrated by the blue boxes showing interquartile range and median accuracy. The case of BW is particularly notable.

\begin{figure}[H]
\centering
\includegraphics[width=\textwidth]{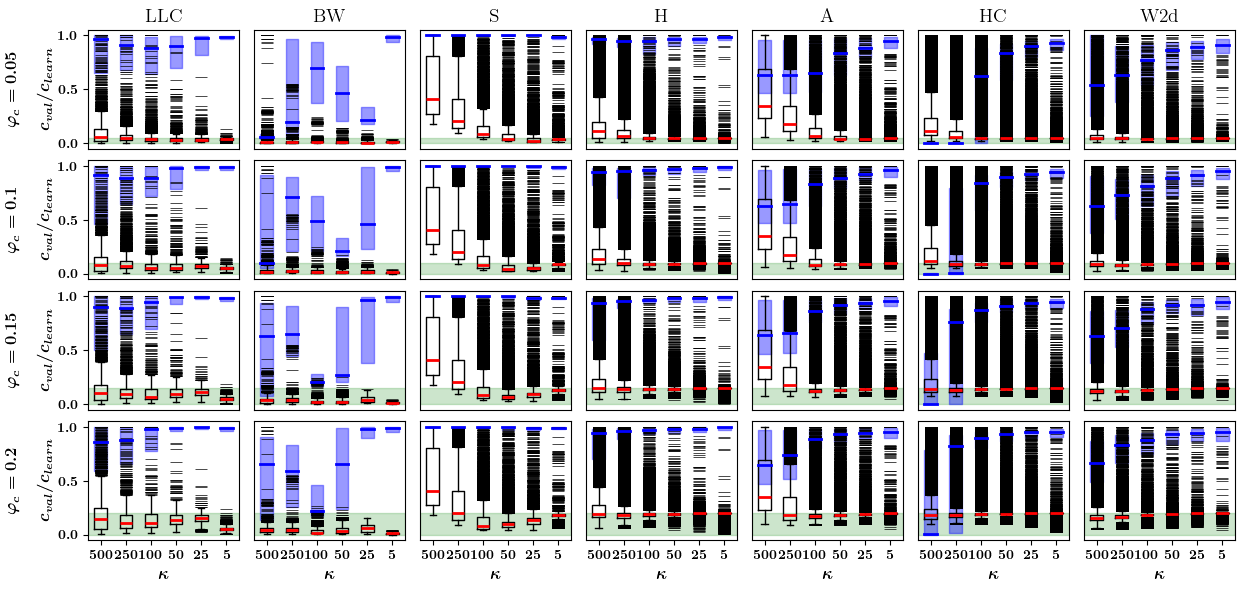}
\vspace{-7mm}
\caption{Boxplots of the validation-to-learning cost ratio for the cost-driven \textit{validation} criterion across the 30 generated learning processes, considering different cost thresholds (rows) and numbers of validation episodes (columns within each plot), for each environment (columns). The green band highlights the range delimited by the corresponding cost threshold, while the blue boxes in the matching columns show the interquartile range and median of the selection accuracy.}\label{FIGURE_cost_test_cost_driven_boxplots}
\vspace{-7mm}
\end{figure}

Figure~\ref{FIGURE_cost_test_cost_driven_boxplots} shows the same analysis for the cost-driven \textit{validation} criterion. In this case, validation cost is much better controlled, with most boxplots lying within the green range delimited by the threshold $\varphi_c$. When surpasses occur, they correspond to cases with a large number of validation episodes $\kappa$. This leads to longer validation periods, making the validation cost estimator (denoted by $\widetilde{c}$ in Algorithm~\ref{ALGORITHM_CostDrivenTest}) less accurate due to larger differences between validated policies, and consequently resulting in higher-than-expected validation costs. In addition, the distribution of selection accuracies for a small number of validation episodes $\kappa=5$ is more stable across environments than in Figure~\ref{FIGURE_cost_test_default_boxplots}.

\vspace{-3mm}

\begin{figure}[H]
\centering
\includegraphics[width=\textwidth]{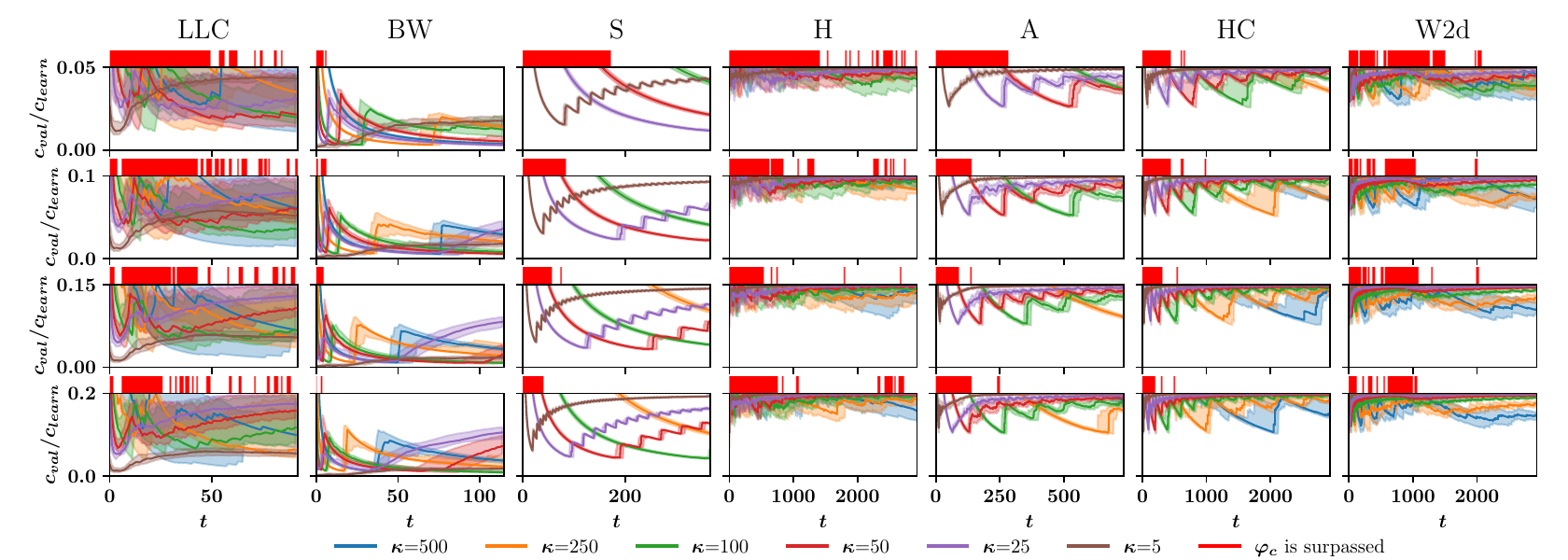}
\caption{Evolution of the validation-to-learning cost ratio for the cost-driven \textit{validation} criterion across the 30 learning processes, considering different cost thresholds (rows) and numbers of validation episodes (different curves), for each environment (columns). Red markers above each plot indicate iterations in which at least one exceedance of the predefined validation-to-learning cost threshold occurred.}\label{FIGURE_cost_cost_driven_test}
\vspace{-5mm}
\end{figure}

Taking a closer look at the cost-driven \textit{validation} criterion, Figure~\ref{FIGURE_cost_cost_driven_test} explicitly shows the evolution of the validation-to-learning cost ratio across iterations for different configurations. Above each plot, red markers indicate iterations where the cost threshold is exceeded. Most of them occur at the beginning of training, where the cost estimator is less accurate due to limited validation observations. However, these cases are outliers, as shown by the boxplots in Figure~\ref{FIGURE_cost_test_cost_driven_boxplots}, whose interquartile range is in most cases fully contained within the threshold green band.

\subsubsection{Benefits of adaptive validation period}

\begin{minipage}[b]{0.48\textwidth}

\begin{figure}[H]
\centering
\includegraphics[width=\textwidth]{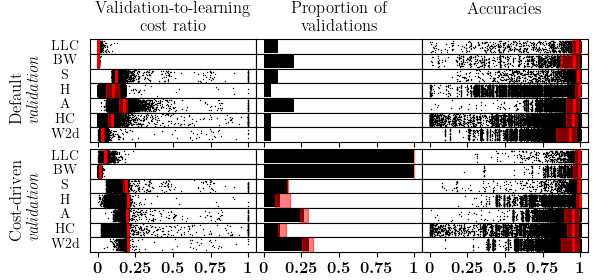}
\caption{In the first and last columns, strip plots of validation-to-learning cost ratios truncated at 1 and accuracies across the 30 policy sequences per environment, together with the interquartile range and median in red. In the central column, the median percentage of validated policies, together with the interquartile range in red.}\label{FIGURE_TestDefaultvsTestCost}
\vspace{-4mm}
\end{figure}

Up to this point, we have shown that the proposed cost-driven \textit{validation} criterion allows, for different values of $\kappa$, a controlled cost under a fixed $\varphi_c$. However, this comes at the expense of a validation frequency that becomes variable not only across environments but also across individual processes within an environment. Figure~\ref{FIGURE_freq_test_cost_driven} shows validation patterns across the 30 sequences used per environment under the selected configuration. Each line corresponds to a seed, where black marks indicate validation iterations. In simpler cases such as LLC and BW, almost every policy is validated, whereas in more complex environments validation becomes more sparse and irregular. These patterns also vary across different sequences within the same environment.

\end{minipage}
\hfill
\begin{minipage}[b]{0.48\textwidth}
\begin{figure}[H]
\centering
\includegraphics[width=\textwidth]{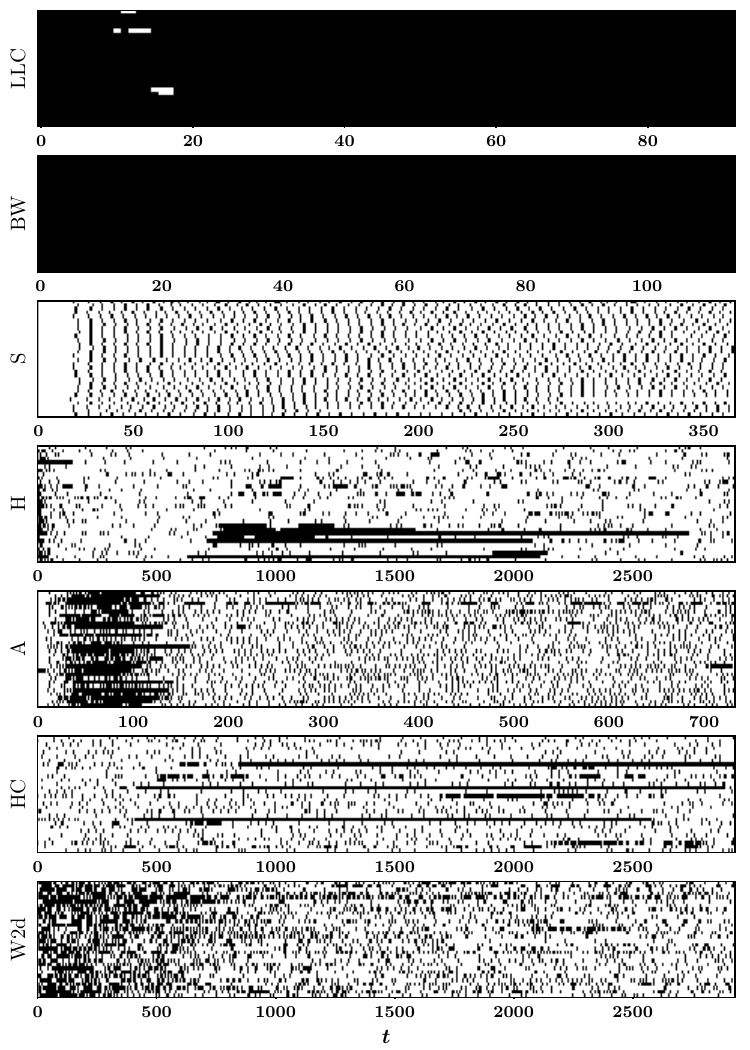}
\caption{Validated policies (marked in black) throughout the 30 policy sequences (one row per sequence in each plot) using the cost-driven \textit{validation} criterion across the different environments.}\label{FIGURE_freq_test_cost_driven}
\end{figure}
\end{minipage}

The adaptive validation exhibits certain advantages compared with the fixed validation period of the default \textit{validation} criterion. In relatively simple environments with lower overall interaction budgets, such as LLC and BW, the default validation period results in too few policy evaluations during training, leading to a decrease in best-policy tracking accuracy. Specifically, high-performing policies are overlooked, despite the fact that validating all policies would not have incurred a substantial computational overhead (compare the first two subrows associated with both the default and cost-driven \textit{validation} criteria in Figure~\ref{FIGURE_TestDefaultvsTestCost}). In contrast, accuracy in these cases is improved by the proposed cost-driven \textit{validation} criterion, which adapts the validation period according to a predefined validation-to-learning cost ratio.

In summary, the cost-driven \textit{validation} criterion proposed in this work defines the validation frequency such that, for a fixed configuration, it yields a similar maximum validation cost across processes. However, for fixed $\kappa$ and a validation period controlled by a cost threshold $\varphi_c$, the resulting \textit{validation} criterion induces irregular validation frequencies both within and across processes. However, a similar effect is observed in the default \textit{validation} criterion, where the frequency is fixed in terms of interaction steps (see the fourth column Table~\ref{TABLE_configurations}). This leads to different numbers of evaluations across environments due to varying interaction lengths~\citep{chan_measuring_2020}, although without preserving any frequency-related invariance across processes, unlike our cost-driven \textit{validation} criterion.

\section{Regions of learning processes}\label{APPENDIX_regions}
\vspace{-7mm}

\begin{figure}[H]
\centering
\includegraphics[width=\textwidth]{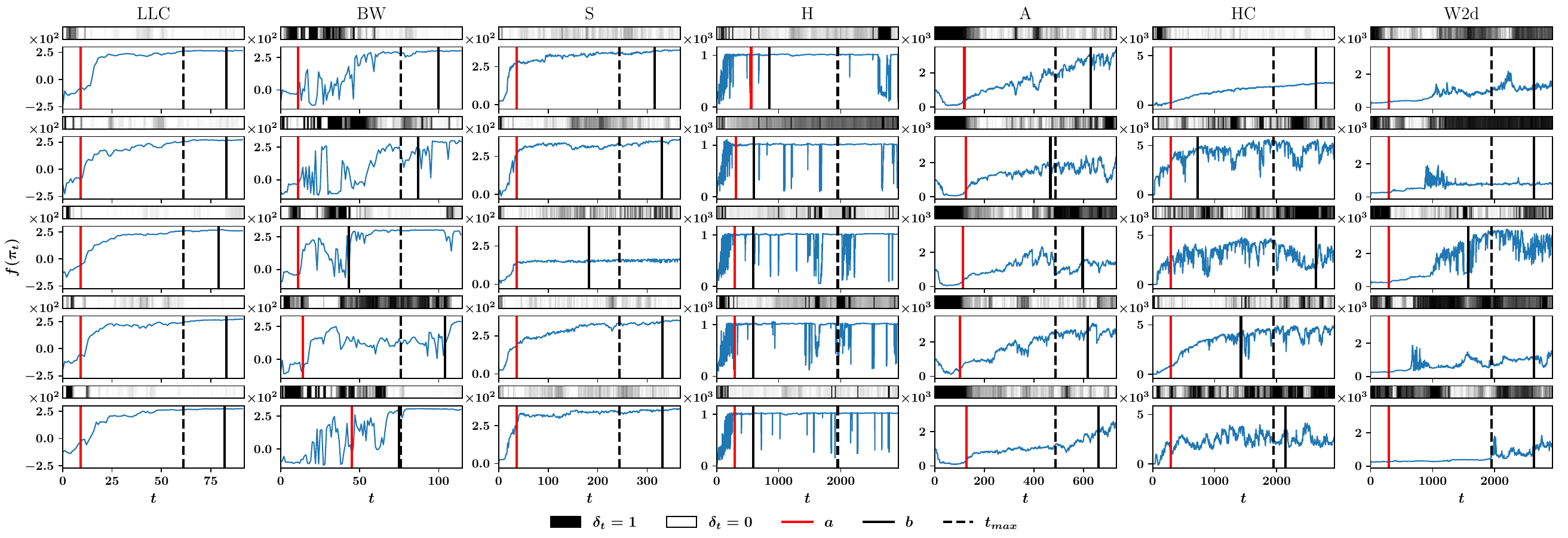}
\caption{Evolution of the ground truth (blue curve) across the generated sequences for each environment (by column) and for 5 different seeds (by row), together with the start of learning (red line) and convergence (black line), which delimit the three regions. Above each curve, the evolution of degradation is shown along the sequence, where darker values indicate higher degradation.}\label{FIGURE_explicit_regions_deg}
\vspace{-5mm}
\end{figure}

\begin{minipage}[b]{0.48\textwidth}
To conduct the experimental analysis in Section~\ref{SECTION_ExperimentalAnalysis}, we use a total of 30 sequence of policies per environment, each with a fixed but environment-dependent length. However, for simplicity, the results are presented in a compact form, summarizing seed variability using statistics such as the interquartile range and characterizing temporal variability of the metrics across three regions. In this appendix, we explicitly show in Figure~\ref{FIGURE_explicit_regions_deg} the characterization of the regions (see Equation~\ref{DEFINITION_regions}) defined over individual sequence per environment that are internally considered in our analysis, together with the evolution of the associated degradation (see Equation~\ref{EQUATION_degradation}) for each case.

\end{minipage}
\hfill
\begin{minipage}[b]{0.48\textwidth}

\begin{table}[H]
\centering
{\footnotesize
\caption{Mean and standard deviation of the percentage of the learning start and convergence start with respect to the total execution horizon for each environment, considering the 30 policy sequences generated in each case.}\label{TABLE_regions}
\begin{tabular}{|c|c|c|}
\hline 
$\mathcal{E}$ & $\frac{a}{T}\cdot 100$ & $\frac{b}{T}\cdot 100$ \\ 
\hline 
LLC & $9.78\pm 0.00$ & $85.33\pm 9.20$ \\ 
\hline 
BW & $14.49\pm 8.28$ & $70.52\pm 21.48$ \\ 
\hline 
S & $9.81\pm 0.00$ & $70.02\pm 23.01$ \\ 
\hline 
H & $12.3\pm 5.01$ & $31.48\pm 23.04$ \\ 
\hline 
A & $16.58\pm 2.50$ & $79.69\pm 10.58$ \\ 
\hline 
HC & $10.79\pm 3.62$ & $68.44\pm 19.74$ \\ 
\hline 
W2d & $10.00\pm 0.00$ & $77.09\pm 16.24$ \\

\hline 
\end{tabular} 
}
\end{table}
\vspace{1mm}

\end{minipage}

Figure~\ref{FIGURE_explicit_regions_deg} shows only 5 seeds per environment, which are sufficient to illustrate how the region definition adapts to different scenarios to correctly characterize the three phases of initialization, learning, and convergence. For instance, while in LLC learning processes are relatively stable with generally low degradation (lighter top bars) and small initialization and convergence regions, in BW we observe more unstable sequences with higher degradation (darker top bars), where the region characterization adapts to each case. In particular, the learning-start limit $a$ tends to occur later when there is higher initial instability (last row), while the convergence-start limit $b$ appears earlier when return improvement stagnates (third row). Table~\ref{TABLE_regions} summarizes the position of the limits $a$ and $b$ for each environment in numerical form, considering all 30 seeds.

\vspace{-2mm}

To define the limits $a$ and $b$ for each learning process, we need to fix the value of the window parameter $w$, on which their definition in Equation~\ref{DEFINITION_regions} depends. The parameter $w$ is part of the post-hoc partitioning procedure and not of the proposed accuracy (see Equation~\ref{EQUATION_accuracy}) and efficiency (see Equation~\ref{EQUATION_efficiency}) diagnostic metrics themselves. It determines the degree of temporal continuity over which the statistical conditions of the sets $A$ and $B$ associated with the limits $a$ and $b$, respectively, are required to hold. Therefore, changing $w$ does not affect the definition or the value of the diagnostic metrics at any individual iteration, but only the resulting region boundaries and, consequently, the aggregation of metric values reported for each region. For our analysis, we set $w$ to 10\% of the total learning horizon. With this value, the resulting regions consistently match the intended interpretation of the initialization, learning, and convergence regions across all $30\times7$ learning processes considered, as illustrated by the sample of $5\times7$ processes shown in Figure~\ref{FIGURE_explicit_regions_deg}.

\section{Selection criteria in early-stopping and resource-allocation techniques}
\label{APPENDIX_EarlyStopping}

In some practical settings, the objective is to obtain the best possible policy for a given task under a fixed computational budget. Such a budget may be large enough to allow multiple independent learning processes to be executed~\citep{echevarrieta_review_2026}, for example by running the same method with different random seeds. However, the learning curves generated by the same method-environment pair can vary substantially across seeds~\citep{cheng_control_2019,henderson_deep_2018-1}, with some runs leading to highly successful policies while others stagnate at relatively poor return levels.

Obtaining a sufficiently large set of learning processes to reliably estimate the best achievable performance of a method can be computationally expensive~\citep{jordan_position_2024,agarwal_deep_2021}. In many situations, the available budget may not be sufficient to fully execute all candidate processes. For this reason, early-stopping and adaptive resource-allocation techniques~\citep{li_hyperband_2018,bohdal_pasha_2023} are particularly valuable, as they discarded early the poorly performing processes, allowing more computational resources to be allocated to more promising ones. A classical approach is successive halving~\citep{jamieson_non-stochastic_2016}, which iteratively discards the worst-performing half of the candidate processes to double the resources allocated to the remaining ones. This procedure is repeated until only a single process remains. Since these decisions are based on comparisons between learning curves, the selection criterion used to represent each learning process may influence the resulting resource-allocation decisions.

\begin{figure}[H]
\centering
\includegraphics[width=\textwidth]{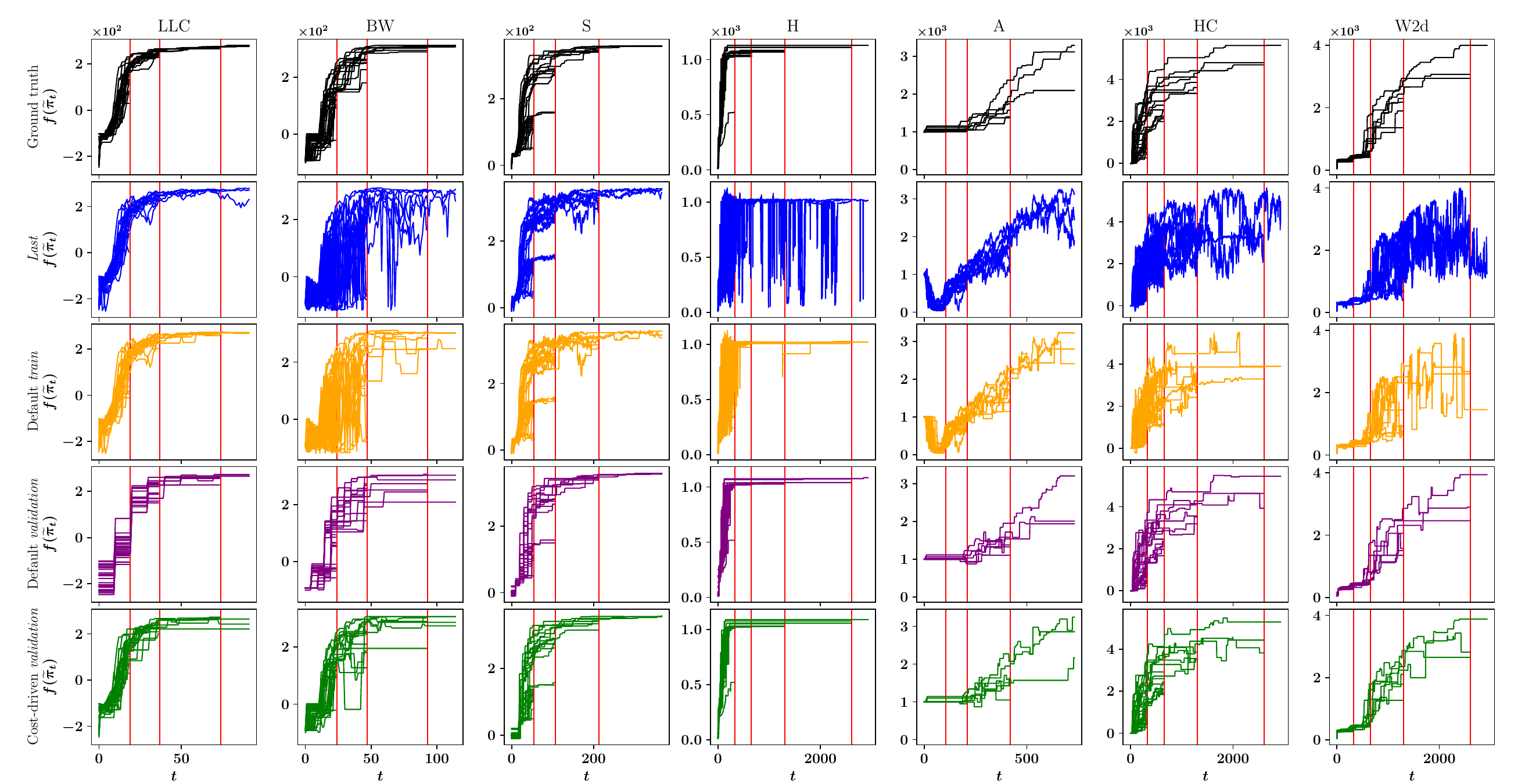}
\caption{For each environment (columns) and selection criterion (rows), application of successive halving to the 30 policy sequences represented as learning curves. Each curve shows the ground-truth return of the best policy identified by the corresponding selection criterion. Red vertical lines indicate successive elimination stages, where half of the surviving processes are discarded according to the estimated return of their currently selected best policies\protect\footnotemark.}\label{FIGURE_SH_LearningCurves}
\end{figure}

In this appendix, we provide a preliminary study of the influence of selection criteria on successive halving. Figure~\ref{FIGURE_SH_LearningCurves} illustrates the application of successive halving to the 30 policy sequences generated for each environment, considering different selection criteria for constructing the corresponding learning curves.

\footnotetext{
For the \textit{last} criterion, which does not rely on return estimates, elimination decisions are based on the default \textit{train} estimator applied to the last policy of each surviving sequences, using the most recent $\kappa=100$ episodes generated during learning.
}

\subsection{Inefficiencies in resource allocation}\label{APPENDIX_IneficcientRA}

Figure~\ref{FIGURE_SH_LearningCurves} reveals differences in the behaviour of successive halving depending on the selection criterion used to represent the learning curves. In particular, the curves associated with the \textit{last} and \textit{train} criteria (second and third rows) tend to occupy lower return levels at later iterations, leaving a larger portion of the area below the diagonal of each plot more densely covered. This suggests that these criteria may spend a larger fraction of the available budget evaluating policy sequences that ultimately achieve relatively poor return.

\begin{figure}[H]
\includegraphics[width=\textwidth]{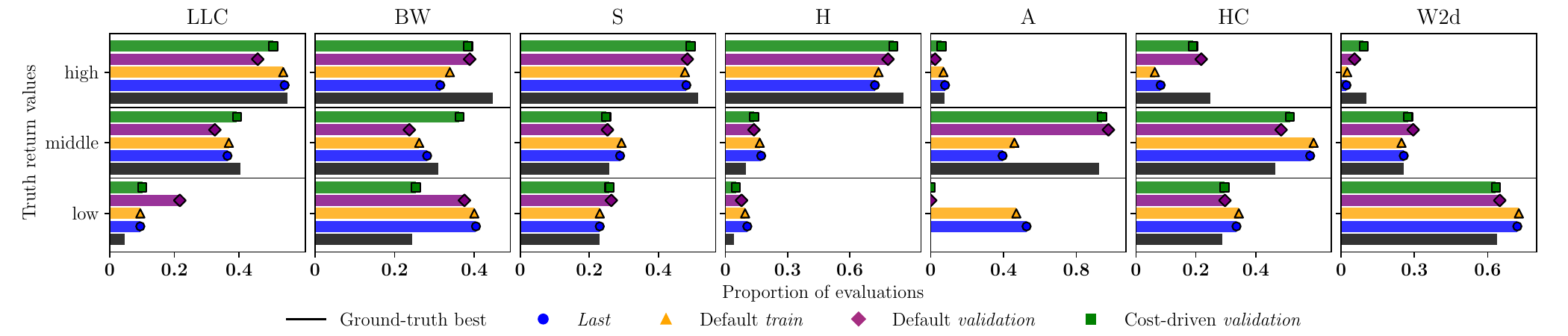}
\caption{Distribution of evaluations allocated by successive halving across different policy return levels for each environment and selection criterion. The observed return range is partitioned into three regions (low, middle and high) using 25\% and 75\% of the maximum observed return as thresholds. Histograms show the proportion of evaluations assigned to policies within each return range. The black bars correspond to the ideal resource-allocation distribution obtained using ground-truth policy return, while the coloured bars show the distributions induced by each selection criterion. Deviations from the reference distribution indicate differences in how computational resources are allocated across low, medium, and high performing policies.}\label{FIGURE_ResourceAllocation}
\vspace{-5mm}
\end{figure}

To further analyze this observation, Figure~\ref{FIGURE_ResourceAllocation} shows the distribution of computational resources allocated across different return levels. The observed return range is divided into three regions, using 25\% and 75\% of the maximum observed return as thresholds. We then compute, for each selection criterion, the distribution of evaluations assigned to policies within each return region and visualize the result using separate histograms. We observe that, the distributions most closely matching the ideal resource allocation correspond to the cost-driven \textit{validation} criterion, where the green bars closely follow the reference distribution shown in black. In contrast, the \textit{last} and \textit{train} criteria exhibit the largest deviations. In these cases, the blue and orange bars tend to exceed the reference in low-return levels, while falling below it in high-return levels. This suggests that the \textit{last} and \textit{train} criteria allocate more resources to poorly performing policies and fewer resources to high-performing ones, resulting in a less efficient allocation of computational resources.

\subsection{Inaccuracies in best-policy tracking}\label{APPENDIX_InaccurateTracking}
The change induced by selection criteria in the allocation of resources across the 30 processes can also affect the tracking of the best policy among processes. During the application of successive halving, at each execution time we have as many candidate best policies as surviving processes. Each of these policies is assigned a return estimate by the selection criterion, from which the policy with the highest estimated return is selected as the best policy learned up to that moment. By comparing the return of this policy with the true best policy, a tracking accuracy for early-stopping techniques applied to a set of processes under a given selection criterion can be defined, similar to how we define selection accuracy over a single learning sequence. For the preliminary experiment, we propose as a definition of \textit{tracking accuracy for a criterion under successive halving} the following adaptation of Equation~\ref{EQUATION_accuracy}

\begin{equation}\label{EQUATION_SH_acc}
\alpha_t^{SH}=\frac{\min \left\{ f\!\left(\underset{\rho,\widetilde{\pi}_t}{\arg\max}\,\widetilde{f}(\widetilde{\pi}_t)\right)-\min_{0,t}, \; \underset{\rho,\pi^*_t}{\max}\, f(\pi^*_t)-\min_{0,t} \right\}}
{\underset{\rho,\pi^*_t}{\max}\, f(\pi^*_t)-\min_{0,t}}
\end{equation}
where 
\[
\text{min}_{0,t}:=\min_\rho \big\{\min\left\{f(\widetilde{\pi}_0),\ldots,f(\widetilde{\pi}_t),f(\pi^*_0),\ldots,f(\pi^*_t)\right\}\big\}
\]

is the minimum return across all learning curves defined by the selection criterion and by the ground truth. In the numerator, the minimum prevents the accuracy from exceeding 1, since when applying successive halving with a given criterion, it may happen that sequences that are discarded early under the ground truth are not discarded, but eventually lead to higher returns.

\begin{figure}[H]
\centering
\includegraphics[width=\textwidth]{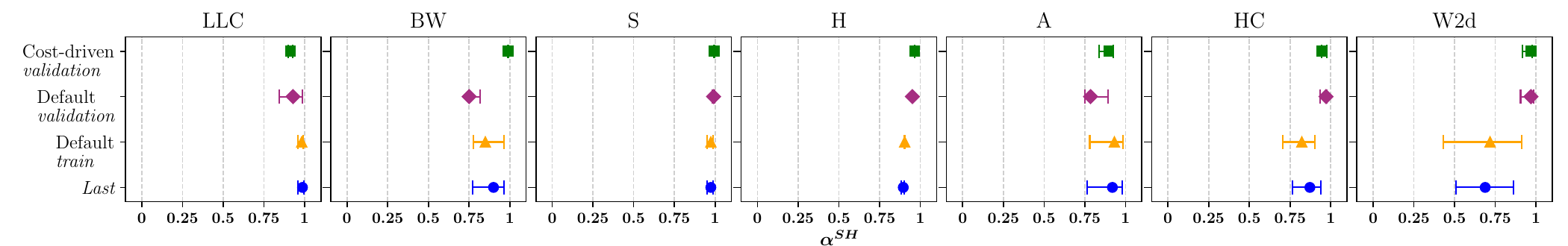}
\caption{Tracking accuracy of successive halving under different selection criteria, computed across 30 policy sequences per environment. For each environment (columns) and selection criterion (rows), segments show the interquartile range and markers indicate median of the tracking accuracy defined in Equation~\ref{EQUATION_SH_acc}.}\label{FIGURE_SH_accuracy}
\end{figure}

In Figure~\ref{FIGURE_SH_accuracy}, we show, for each environment and criterion pair, the distribution of tracking accuracy using the interquartile range represented by the segment and the median indicated by the marker. We observe that the tracking accuracy of the \textit{last} and \textit{train} criteria is lower in more complex environments such as HC and W2d, with median values below 0.75 in the latter case. In contrast, validation-based criteria achieve higher tracking accuracy, with narrow interquartile ranges and values close to 1 in these environments. It is worth noting that the cost-driven \textit{validation} criterion maintains consistently high tracking accuracy across environments, unlike default \textit{validation}, which shows drops in accuracy in certain environments such as BW and A, with values close to 0.75. These observations suggest that the cost-driven \textit{validation} criterion could accurately track the best solution in successive halving processes.

\subsection{Finding trade-off validation-to-learning cost ratio threshold}\label{APPENDIX_CostThresholdTradeoff}
\vspace{-5mm}
\begin{figure}[H]
\centering
\includegraphics[width=\textwidth]{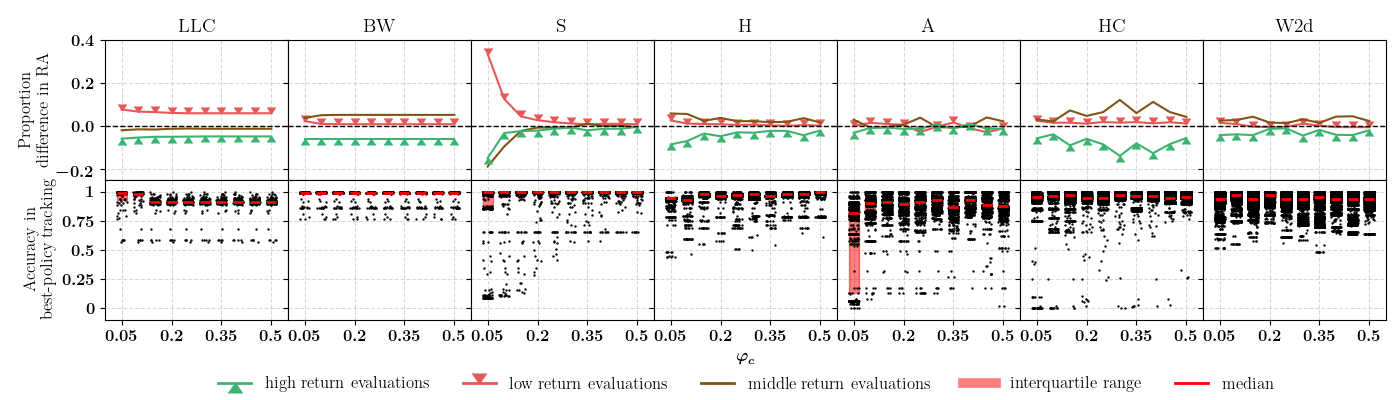}
\caption{Trade-off analysis between Resource Allocation (RA) efficiency and best-policy tracking accuracy as a function of the cost threshold $\varphi_c$. In the top row, differences in the distribution of evaluations across low, medium, and high return ranges after applying successive halving, measured relative to the ground-truth allocation. In the bottom row, the distribution of tracking accuracy for each value of $\varphi_c$, showing interquartile range and median in red.}\label{FIGURE_SH_tradeoff}
\vspace{-5mm}
\end{figure}

The cost-driven \textit{validation} criterion has shown strong results across the different analyses conducted in this paper. In particular, it has demonstrated consistent accuracy and efficiency across environments and training regions that induce policy sequences with varying levels of degradation. However, this criterion depends on the parameters $\kappa$ and $\varphi_c$, which determine the number of validation episodes and the validation-to-learning cost threshold used to adapt the validation frequency during learning. We have conducted our analyses using the configuration $\kappa=5$ and $\varphi_c=0.2$, under which we generally observe high selection accuracy across environments and regions (see Appendix~\ref{APPENDIX_Configurations_Experiments}). As a final experiment, in Figure~\ref{FIGURE_SH_tradeoff}, we analyze the influence of the parameter $\varphi_c$ on the trade-off between efficient resource allocation in successive halving and accurate tracking of the best policy.

Concretely, in the first row of Figure~\ref{FIGURE_SH_tradeoff} we show, for each value of $\varphi_c$, the percentage difference in evaluations allocated to high, medium, and low return ranges after applying successive halving. These values illustrate the deviations between the green and black bars in Figure~\ref{FIGURE_ResourceAllocation}. Then, in the second row, for each value of $\varphi_c$, a strip plot shows the tracking accuracies obtained using Equation~\ref{EQUATION_SH_acc}, together with the interquartile range and median in red.

In general, we observe that for different values of $\varphi_c$, fewer evaluations tend to be allocated to high-return regions (the green curve lies below the horizontal zero line), while more evaluations are allocated to low-return levels (the pink curve lies above it), consistent with the general trends observed in Figure~\ref{FIGURE_ResourceAllocation}. However, in most cases these curves remain close to the horizontal baseline, suggesting that the deviations from the ground-truth resource allocation are not large. Additionally, although no strong trend is observed across values of $\varphi_c$ in most environments, the case of S is notable. In this environment, for values of $\varphi_c$ below 0.2, resource allocation becomes significantly less efficient, while inaccuracies in best-policy tracking also begin to appear. The pink and green curves diverge, while cases with tracking accuracies below 0.25 become more frequent. This suggests that, in this setting, using a threshold above 0.2 does not provide significant gains in efficiency or accuracy, while smaller values may lead to inefficiencies and inaccuracies. Therefore, the value $\varphi_c=0.2$ appears to define a good trade-off. However, in other settings where results are less sensitive to $\varphi_c$, a smaller value could achieve similar accuracy and efficiency while reducing validation cost, leaving more budget available to execute additional processes.

\section{Construction of the figures for the experimental analysis }
\label{APPENDIX_FigureConstruction}

A method $\mathcal{M}$, an environment $\mathcal{E}$, a random seed $\rho$, and an execution time $t$ determine a learning process, which we denoted as  $\mathcal{M}(\mathcal{E},\rho,t)\rightarrow \widetilde{\pi}_t$ (see Section~\ref{SECTION_TheoreticalFramework} and Algorithm~\ref{ALGORITHM}). During this process, a sequence of policies $\Pi_t$ is generated and the policy $\widetilde{\pi}_t$ returned as learned by a selection criterion  belongs to this sequence.

Our experimental analysis is based on a single on-policy Proximal Policy Optimization~\citep{schulman_proximal_2017-1} method and seven Gymnasium~\citep{towers_gymnasium_2025} environments
$\mathcal{E}\in \{\text{LLC},\text{BW},\text{S},\text{H},\text{A},\text{HC},\text{W2d}\}$, as we described in Section~\ref{SECTION_ExperimentalFramework}. For each environment, we generate 30 policy sequences by considering 30 different random seeds and a maximum execution time $T$. Each sequence of $T$ policies is characterized through the three regions defined in Equation~\ref{DEFINITION_regions}. This allows us to partition a complete sequence of $T$ policies into subsequences of length $t$, where $t\in R$ being $R\in\left\{[0,a],[a,b],[b,T]\right\}$ one of the intervals defining the regions. More formally, given a sequence $\Pi_T=(\pi_0,...,\pi_T)$, the region $R$ have associated the following set of subsequences:
\begin{equation}
\Pi_{\mathrm{sub}}
=
\{\Pi_t: t\in R\}.
\end{equation}

The figures presented throughout Section~\ref{SECTION_ExperimentalAnalysis} are constructed from the values of the proposed metrics computed over these subsequences. In this appendix, we complement the descriptions provided in Section~\ref{SECTION_ExperimentalAnalysis} by explicitly detailing how each figure is constructed.

\subsection{Figures based on the complete set of subsequences}\label{APPENDIX_FigureConstruction_Complete}

The metrics considered throughout the analysis are degradation, selection accuracy, and selection efficiency, defined in Equations~\ref{EQUATION_degradation}, \ref{EQUATION_accuracy} and \ref{EQUATION_efficiency}, respectively. The corresponding samples used for each figure are described below:

\paragraph{First plot in Figure~\ref{FIGURE_analysis_deg_acc_eff}.}

Given an environment and a learning region, we consider the sample of degradation values associated with all subsequences whose lengths belong to the interval $R$, formally defined as

\begin{equation}
X_\delta
=
\bigcup_{\rho\in\{1,\ldots,30\}}
\{\delta_t : t\in R\}.
\end{equation}

This sample is used to construct a histogram with 10 bins over the degradation range $[0,1]$. Instead of representing frequencies through bar heights, darker colors indicate a larger number of degradation observations within the corresponding interval. The sample median is additionally represented by a red vertical line.

\paragraph{Second and third plots in Figure~\ref{FIGURE_analysis_deg_acc_eff}.}

Given an environment, a learning region, and a selection criterion, we consider the samples of selection accuracy and selection efficiency associated with all subsequences whose lengths belong to the interval $R$, formally defined as

\begin{equation}
X_\alpha
=
\bigcup_{\rho\in\{1,\ldots,30\}}
\{\alpha_t : t\in R\},
\end{equation}

\begin{equation}
X_\varepsilon
=
\bigcup_{\rho\in\{1,\ldots,30\}}
\{\varepsilon_t : t\in R\}.
\end{equation}

The distributions of accuracy and efficiency are represented through the interquartile range together with the median of these samples.

\paragraph{Figure~\ref{FIGURE_ScatterPlots_deg_acc_eff}.}

Given an environment, for each pair consisting of a selection criterion and a learning region, we represent the point $\bigl(\mathrm{Median}(X_\delta),
\mathrm{Median}(X_\alpha)\bigr)$ to study the relationship between degradation and selection accuracy, and $\bigl(\mathrm{Median}(X_\delta),\mathrm{Median}(X_\varepsilon)\bigr)$ to study the relationship between degradation and selection efficiency.

\subsection{Figures based on pairwise criterion comparisons}\label{APPENDIX_FigureConstruction_Pairwise}

For the comparative analysis of selection criteria presented in Section~\ref{SECTION_ExperimentalAnalysis}, the previous samples are restricted to the subsequences for which one criterion outperforms another. Specifically, given an environment, a learning region, and two selection criteria, we focus on the subsequences $\Pi_t\in\Pi_{\text{sub}}$ such that $t$ belongs to

\begin{equation}
T_{1,2}
=
\bigcup_{\rho\in\{1,\ldots,30\}}
\left\{
t\in R :
f\left(\widetilde{\pi}_t^{\,1}\right)
>
f\left(\widetilde{\pi}_t^{\,2}\right)
\right\},
\end{equation}

where $\widetilde{\pi}_t^{\,1}$ and $\widetilde{\pi}_t^{\,2}$ denote the policies selected by the first and second criteria, respectively. The set $T_{1,2}$ therefore represents all subsequences for which the first criterion selects a policy with higher ground-truth return than the second criterion.

The corresponding figures are constructed as follows:

\paragraph{Figure~\ref{FIGURE_comparison_times}.}

Given an environment, two selection criteria, and a learning region, we compare the cardinals $|T_{1,2}|$ and $|T_{2,1}|$ with respect to the total number of subsequences, including in the total also ties.

\paragraph{Figure~\ref{FIGURE_comparison_deg}.}

Given an environment, two selection criteria, and a learning region, we consider the degradation samples associated with the subsequences for which each criterion outperforms the other:

\begin{equation}
X_\delta^{1}
=
\bigcup_{\rho\in\{1,\ldots,30\}}
\{\delta_t : t\in T_{1,2}\},
\end{equation}

\begin{equation}
X_\delta^{2}
=
\bigcup_{\rho\in\{1,\ldots,30\}}
\{\delta_t : t\in T_{2,1}\}.
\end{equation}

These samples are used to construct two 10-bin histograms following the same procedure described for Figure~\ref{FIGURE_analysis_deg_acc_eff}. Finally, both histograms are jointly normalized using the maximum frequency observed across the two samples, and the normalized frequencies are represented through color intensity. Additionally, the median of each subset is shown with a marker.

\end{document}